\documentclass[10pt,twocolumn,letterpaper]{article}
\usepackage{dingbat}
\usepackage{graphicx}
\usepackage{subcaption}
\usepackage{subcaption}
\usepackage{algorithm}
\usepackage{pifont}
\usepackage{algpseudocode}
\usepackage{amsmath}
\usepackage{soul}
\usepackage{booktabs}   
\usepackage{adjustbox}  
\usepackage{multirow}   
\usepackage[table]{xcolor}
\newcolumntype{C}{>{\centering\arraybackslash}X}
\usepackage{wacv}              

\definecolor{wacvblue}{rgb}{0.21,0.49,0.74}
\definecolor{forestgreen}{rgb}{0.133,0.549,0.133}
\newcommand{\cmark}{\textcolor{forestgreen!80!black}{\ding{51}}}
\sethlcolor{wacvblue!10}

\newcommand{\ours}{\texttt{PROBE}\xspace}

\usepackage[pagebackref,breaklinks,colorlinks,allcolors=wacvblue]{hyperref}

\def\wacvPaperID{2094} 
\def\confName{WACV}
\def\confYear{2026}

\title{\ours: Self-Supervised Visual Prompting for\\Cross-Domain Road Damage Detection}

\author{
\textbf{Xi Xiao}\textsuperscript{1,6,*} \quad
\textbf{Zhuxuanzi Wang}\textsuperscript{2,*} \quad
\textbf{Mingqiao Mo}\textsuperscript{2,*} \quad
\textbf{Chen Liu}\textsuperscript{3}\quad
\textbf{Chenrui Ma}\textsuperscript{4}\\
\textbf{Yanshu Li}\textsuperscript{5}\quad
\textbf{Smita Krishnaswamy}\textsuperscript{3}\quad
\textbf{Xiao Wang}\textsuperscript{6}\quad
\textbf{Tianyang Wang}\textsuperscript{1}\textsuperscript{\dag}\\[3mm]
\textsuperscript{1}University of Alabama at Birmingham, Birmingham, AL, USA\\
\textsuperscript{2}Cornell University, Ithaca, NY, USA\\
\textsuperscript{3}Yale University, New Haven, CT, USA\\
\textsuperscript{4}University of California Irvine, Irvine, CA, USA\\
\textsuperscript{5}Brown University, Rhode Island, USA\\
\textsuperscript{6}Oak Ridge National Laboratory, Oak Ridge, TN, USA\\
\small\textsuperscript{*}Equal contribution \\[2mm]
{\small \textsuperscript{\dag}Corresponding authors: \texttt{wangx@ornl.gov, tw2@uab.edu}}
}

\begin{document}
\maketitle
\begin{abstract}
The deployment of automated pavement defect detection is often hindered by poor cross-domain generalization. Supervised detectors achieve strong in-domain accuracy but require costly re-annotation for new environments, while standard self-supervised methods capture generic features and remain vulnerable to domain shift. We propose \ours, a self-supervised framework that \emph{visually probes} target domains without labels. \ours introduces a Self-supervised Prompt Enhancement Module (SPEM), which derives defect-aware prompts from unlabeled target data to guide a frozen ViT backbone, and a Domain-Aware Prompt Alignment (DAPA) objective, which aligns prompt-conditioned source and target representations. Experiments on four challenging benchmarks show that \ours consistently outperforms strong supervised, self-supervised, and adaptation baselines, achieving robust zero-shot transfer, improved resilience to domain variations, and high data efficiency in few-shot adaptation. These results highlight self-supervised prompting as a practical direction for building scalable and adaptive visual inspection systems. Source code is publicly available: https://github.com/xixiaouab/PROBE/tree/main
\end{abstract}
    
\section{Introduction}
\label{sec:intro}

Automated pavement defect detection is critical for road safety and infrastructure maintenance, as undetected cracks and potholes can escalate into severe hazards and costly repairs. Current practice relies on supervised detectors such as Faster R-CNN and the YOLO family~\cite{ren2015faster,yolov5}. While these models achieve high accuracy within the training domain, they require exhaustive re-annotation to sustain performance in new environments and degrade sharply under domain shifts caused by variations in pavement materials, lighting, or weather~\cite{ganin2016dann,saito2018mcd}. This dependence on large-scale labeling severely limits their scalability in real-world deployments.

Self-supervised learning (SSL) has emerged as a promising alternative by learning transferable representations from unlabeled data~\cite{chen2020simclr,he2020moco,chen2021simsiam}. Although SSL reduces reliance on labeled data, its learned features are often too generic and lack explicit mechanisms for cross-domain alignment, leading to performance drops when applied to specialized tasks like defect detection~\cite{xu2021cdtrans}.

In parallel, Visual Prompt Tuning (VPT) offers a parameter-efficient way to adapt large Vision Transformers (ViTs) by inserting a small set of learnable tokens~\cite{dosovitskiy2020vit,jia2022vpt}. VPT and its extensions have shown success in supervised settings~\cite{khattak2023maple,han2023e2vpt}, but their application to domain adaptation remains limited. Prompts are typically optimized with labels and randomly initialized, failing to leverage the semantic structure embedded in unlabeled target-domain data. This gap motivates our key question: \emph{Can we design prompts that are self-supervised, domain-aware, and specialized for defect detection?}

We address this challenge in the context of \emph{single-source $\rightarrow$ single-target unsupervised domain adaptation (UDA)} for road-damage detection in a closed-set setting. During pre-training, the model has access to unlabeled source and target images, but no target labels. A lightweight detection head is subsequently trained using a small labeled subset of the source domain, with optional few-shot experiments on the target domain reported separately.

We present \ours, a prompt-enhanced self-supervised framework designed for robust cross-domain pavement defect detection. \ours integrates two complementary modules: (1) a \textbf{Self-supervised Prompt Enhancement Module (SPEM)} that derives visual prototypes from unlabeled target images and projects them into prompts that condition a frozen ViT backbone, and (2) a \textbf{Domain-Aware Prompt Alignment (DAPA)} objective that aligns prompt-enhanced source and target features via a linear-kernel MMD loss. With the backbone frozen, only lightweight modules and the detection head are trained, yielding a practical and efficient pipeline.

Our main contributions are as follows:
\begin{itemize}
    \item We propose a \emph{target-aware self-supervised prompting} mechanism that generates prompts from unlabeled target-domain images, enabling the model to capture fine-grained, defect-specific semantics beyond generic SSL representations.
    \item We introduce a \emph{domain-aware alignment strategy} (DAPA) that operates in the prompt-enhanced feature space, explicitly reducing domain discrepancy and improving cross-domain generalization.
    \item We demonstrate through extensive experiments that \ours consistently outperforms strong SSL, UDA, and prompt-tuning baselines across multiple benchmarks, while being highly parameter-efficient with a frozen backbone and lightweight adaptation modules.
\end{itemize}

Taken together, \ours shows that prompts distilled from unlabeled target data can serve as semantic anchors for both specialization and alignment, paving the way for practical self-supervised adaptation in real-world infrastructure inspection.

\section{Related Work}
\label{sec:related}

\subsection{Supervised Pavement Defect Detection}
The prevailing paradigm for pavement defect detection has been supervised learning. Early work employed patch-level CNN classifiers~\cite{zhang2016crack}, followed by modern object detection architectures such as Faster R-CNN~\cite{ren2015faster}, the YOLO family~\cite{yolov5}, and attention-based variants~\cite{pan2023mgd,xiao2025td,li2024cycle,xiao2025roadbench,li2025mgd,lan2025visual,lan2025acam,9904187}. These methods achieve strong accuracy when sufficient annotations are available and have become widely adopted in practice. However, their reliance on domain-specific labels creates a critical bottleneck: models trained in one region often fail to generalize to new environments with different road materials, lighting, or weather. This domain brittleness makes supervised detection difficult to scale without costly and repeated re-annotation~\cite{ganin2016dann}.

\subsection{Self-Supervised Representation Learning}
Self-supervised learning (SSL) provides a powerful alternative by learning transferable representations from unlabeled data. Contrastive and Siamese frameworks such as SimCLR~\cite{chen2020simclr}, MoCo~\cite{he2020moco}, and SimSiam~\cite{chen2021simsiam} have significantly reduced the dependency on large labeled corpora. Yet, applying generic SSL representations directly to defect detection exposes two persistent shortcomings. First, these representations often lack the fine-grained, task-specific semantics necessary to capture subtle cracks or potholes. Second, standard SSL does not address domain discrepancy, leaving models vulnerable to performance drops under domain shift~\cite{xu2021cdtrans}. Recent efforts on defect-aware SSL show promise, but explicit mechanisms for cross-domain adaptation remain underexplored.

\subsection{Domain Adaptation for Object Detection}
Unsupervised domain adaptation (UDA) has been widely studied for object detection to improve robustness across distributions. Early approaches used adversarial alignment~\cite{ganin2016dann, saito2018mcd}, classifier discrepancy~\cite{saito2018mcd}, or feature-level regularization~\cite{tzeng2017adda, he2019multi}. More recent work has adapted transformers for UDA, such as CDTrans~\cite{xu2021cdtrans}, and explored source-free adaptation where only the source-trained model is available at adaptation time~\cite{liang2020weif}. These strategies demonstrate the importance of aligning cross-domain features but often operate on global statistics, which may dilute fine-grained defect cues that are critical in road inspection.

\subsection{Visual Prompt Tuning}
Visual prompt tuning (VPT)~\cite{jia2022vpt,xiao2025focus,xiao2025prompt,xiao2025via,xiao2025visual,han2024facing,zhangdpcore} has recently emerged as a parameter-efficient alternative to full fine-tuning of Vision Transformers (ViT)~\cite{dosovitskiy2020vit}. By inserting a small number of learnable tokens into a frozen backbone, VPT enables efficient adaptation and has shown strong performance in supervised scenarios~\cite{khattak2023maple,han2023e2vpt}. However, its application to domain adaptation remains limited. Existing methods optimize prompts using labels and typically initialize them randomly, overlooking the semantic structure in unlabeled target data. To our knowledge, no prior work has explored generating and adapting prompts in a fully self-supervised manner to simultaneously capture defect-specific semantics and enhance cross-domain robustness. This gap motivates our proposed framework, which leverages unlabeled target-domain imagery to derive task-aware prompts and explicitly align them across domains.

\section{Methodology}
\label{sec:method}

\begin{figure*}[t]
  \centering
  \includegraphics[width=\textwidth]{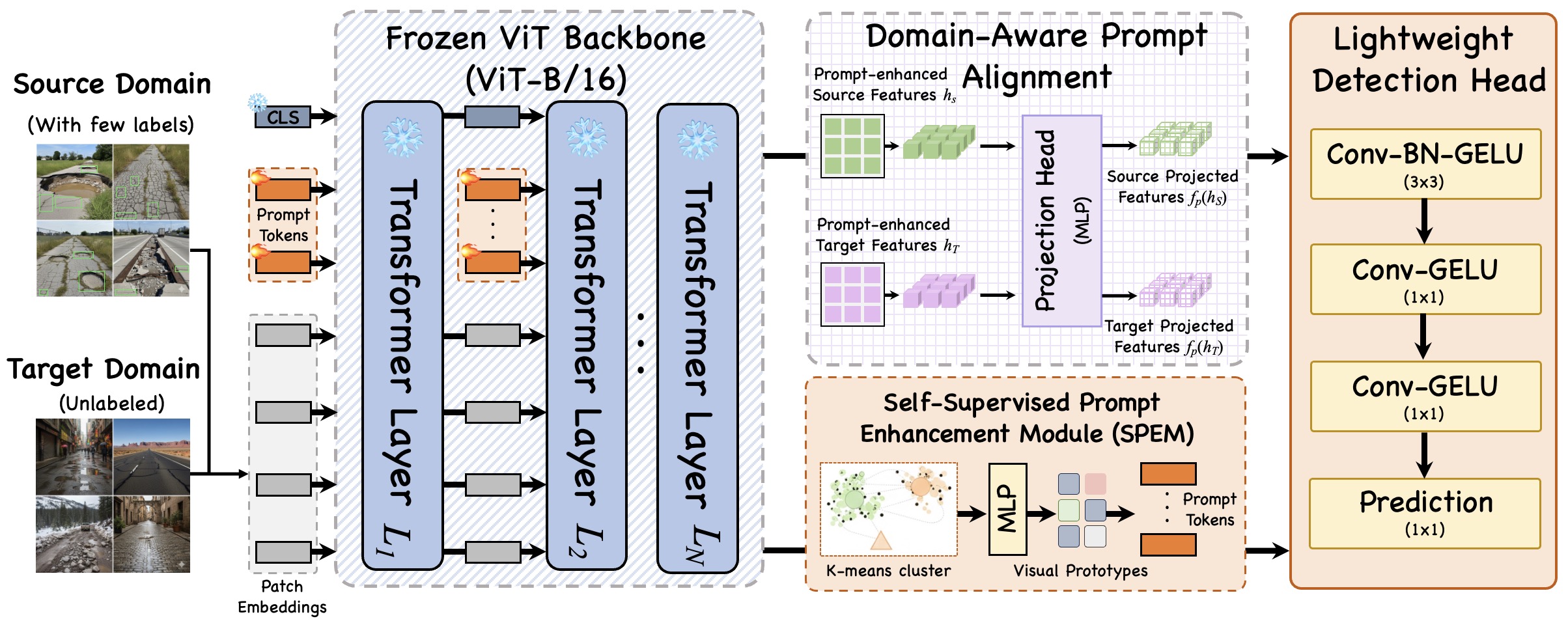}
  \vspace{-2mm}
  \caption{\textbf{Overview of \textsc{PROBE}.}
  The framework targets cross-domain road damage detection with a frozen ViT-B/16 backbone and two key modules:
  (i) \textbf{SPEM} (bottom-right) converts unlabeled target-domain patch embeddings into \emph{visual prototypes} via PCA + K-means and projects them with a shallow MLP into \emph{prompt tokens}, which are injected at shallow and mid transformer layers to emphasize defect-relevant semantics; 
  (ii) \textbf{DAPA} (top-right) aligns \emph{prompt-enhanced} source/target features using a linear-kernel MMD in the prompt-conditioned space.
  A lightweight detection head (right) with two conv blocks and a $1{\times}1$ prediction layer is trained on a small labeled source subset (few-shot target labels optional). 
  }
  \label{fig:probe_overview}
  \vspace{-3mm}
\end{figure*}

We present \ours, a prompt-enhanced self-supervised transfer framework for robust cross-domain pavement defect detection. As shown in Figure~\ref{fig:probe_overview}, the architecture comprises: (i) a \textbf{Self-supervised Prompt Enhancement Module (SPEM)} that derives \emph{target-aware} visual prompts from unlabeled target data, (ii) a \textbf{Domain-Aware Prompt Alignment (DAPA)} objective that aligns \emph{prompt-enhanced} source/target features, and (iii) a lightweight downstream \emph{detection head} trained with limited source labels while the ViT backbone remains frozen.

\subsection{Problem Setup and Notation}
\label{sec:setup}

We study \emph{single-source $\rightarrow$ single-target} \emph{unsupervised domain adaptation (UDA)} in a closed-set regime. During self-supervised pre-training, the model has access to unlabeled images from the source and target domains,
$X^s=\{\mathbf{x}^s\}$ and $X^t=\{\mathbf{x}^t\}$, but \emph{no} target labels are used. In the subsequent detection stage, a small labeled subset from the source domain is used to train a lightweight detector; optional few-shot target-label experiments are reported separately.

Our backbone is a frozen ViT~\cite{dosovitskiy2020vit} with $L$ transformer layers and embedding dimension $D$ (ViT-B/16: $D{=}768$). For an image $\mathbf{x}$, the patchifying projection yields
$\mathbf{z}^{(0)} \!\in\! \mathbb{R}^{N\times D}$ ($N$ tokens), and the $l$-th layer computes
\begin{equation}
\mathbf{z}^{(l)} = \mathrm{TransformerLayer}^{(l)}\!\big(\mathbf{z}^{(l-1)}\big),\quad l=1,\dots,L.
\end{equation}
We denote the final representation as $\mathbf{h}=g(\mathbf{z}^{(L)})$, where $g(\cdot)$ may select a token (e.g., \texttt{[CLS]}) or a pooled feature.

\subsection{Self-Supervised Prompt Enhancement Module (SPEM)}
\label{sec:spem}

\paragraph{Motivation}
Generic SSL features tend to under-emphasize subtle, localized defect cues. SPEM turns unlabeled \emph{target} images into \emph{visual prototypes} and projects them into learnable prompts that condition the frozen ViT at selected layers, steering representation learning toward defect-relevant patterns.

\paragraph{Prototype discovery (global over $X^t$)}
Given the target set $X^t$, we extract patch embeddings with the frozen ViT and aggregate them across the whole target domain. To improve robustness and efficiency, we first reduce dimensionality by PCA from $D$ to $d'$ (e.g., $d'{=}50$). We then run K-means to obtain $K$ centroids (visual prototypes)
\[
\mathcal{C}=\{\mathbf{c}_k\}_{k=1}^{K},\quad \mathbf{c}_k\in\mathbb{R}^{d'}.
\]

\paragraph{Prompt projection and injection}
A shallow projection head (two-layer MLP with GELU) maps prototypes back to the ViT embedding space:
\begin{equation}
\label{eq:proj}
\mathbf{P}^t = \mathrm{MLP}_{\theta_p}(\mathcal{C}) \in \mathbb{R}^{K\times D}.
\end{equation}
We inject $\mathbf{P}^t$ at selected layers (\emph{Shallow+Mid}, e.g., input $L{=}0$ and an intermediate layer $L{=}6$): at an injection layer, the token sequence is augmented by concatenation
\begin{equation}
\label{eq:concat}
\tilde{\mathbf{z}}^{(l-1)}=\big[\mathbf{P}^t;\ \mathbf{z}^{(l-1)}\big]\in\mathbb{R}^{(K+N)\times D},
\end{equation}
and forwarded to layer $l$. The backbone remains frozen; only $\theta_p$ and heads are trainable.

\paragraph{Prompt consistency loss}
To ensure that prompts encode coherent semantics rather than noise, we encourage each image representation to be closer to \emph{its own} prompts than to others. For a target image $\mathbf{x}_i^t$ with final representation $\mathbf{h}_i^t$ and its prompt set mean $\bar{\mathbf{p}}_i^t=\frac{1}{K}\sum_{k=1}^{K}\mathbf{p}_{i,k}^t$, we minimize a temperature-scaled InfoNCE-style objective:
\begin{equation}
\label{eq:l_prompt}
\mathcal{L}_{\mathrm{prompt}}
= -\sum_{i=1}^{B}
\log
\frac{\exp\big(\mathrm{sim}(\mathbf{h}_i^t,\bar{\mathbf{p}}_i^t)/\tau\big)}
{\sum_{j=1}^{B}\exp\big(\mathrm{sim}(\mathbf{h}_i^t,\bar{\mathbf{p}}_j^t)/\tau\big)},
\end{equation}
where $\mathrm{sim}(\cdot,\cdot)$ is cosine similarity and $\tau\!>\!0$ is a temperature.

\subsection{Domain-Aware Prompt Alignment (DAPA)}
\label{sec:dapa}

\paragraph{Motivation}
Aligning \emph{global} features may dilute sparse defect cues. DAPA operates \emph{in the prompt-enhanced space}, aligning distributions of source/target representations after prompt conditioning.

\paragraph{Linear-kernel MMD alignment}
Let $\mathbf{h}^s$ and $\mathbf{h}^t$ be prompt-enhanced representations from source and target images, and $f_p(\cdot)$ a small projection head. We use a linear-kernel MMD objective, equivalent to the squared Euclidean distance between empirical means:
\begin{equation}
\label{eq:l_dapa}
\mathcal{L}_{\mathrm{DAPA}}=
\left\|
\mathbb{E}_{\mathbf{x}^s\sim X^s}\!\big[f_p(\mathbf{h}^s)\big]
-
\mathbb{E}_{\mathbf{x}^t\sim X^t}\!\big[f_p(\mathbf{h}^t)\big]
\right\|_2^2.
\end{equation}
This provides a simple, efficient proxy for reducing cross-domain discrepancy in the prompt-enhanced space.

\subsection{Self-Supervised Objective and Optimization}
\label{sec:ssl_obj}

We adopt a SimSiam-style SSL loss $\mathcal{L}_{\mathrm{ssl}}$~\cite{chen2021simsiam} (no negatives, stop-gradient) and jointly optimize with the prompt consistency and DAPA losses:
\begin{equation}
\label{eq:l_total}
\mathcal{L}_{\mathrm{total}}=
\mathcal{L}_{\mathrm{ssl}}
+\lambda_{1}\,\mathcal{L}_{\mathrm{prompt}}
+\lambda_{2}\,\mathcal{L}_{\mathrm{DAPA}},
\end{equation}
where $\lambda_{1},\lambda_{2}\!>\!0$ balance specialization (SPEM) and alignment (DAPA). Trainable parameters include the prompt projector $\theta_p$, the SSL projector/predictor heads, and $f_p$; the ViT backbone is frozen throughout pre-training.

\subsection{Downstream Detection with a Lightweight Head}
\label{sec:det_head}

After self-supervised pre-training, we freeze the prompt-enhanced ViT and train a lightweight convolutional detection head $g_{\phi}(\cdot)$ on a small labeled subset of the \emph{source} domain. Given the final patch tokens $\mathbf{z}^{(L)}\!\in\!\mathbb{R}^{N\times D}$, we reshape them into a feature map (e.g., $14{\times}14{\times}768$ for 224$\times$224 inputs) and apply a three-stage head:
\begin{align}
&\text{Conv-BN-GELU}(3{\times}3)\ \rightarrow\ 14{\times}14{\times}384, \nonumber\\
&\text{Conv-GELU}(1{\times}1)\ \rightarrow\ 14{\times}14{\times}128, \nonumber\\
&\text{Prediction}(1{\times}1)\ \rightarrow\ 14{\times}14{\times}(C{+}4), \nonumber
\end{align}
where $C$ is the number of defect classes and $4$ parameterizes bounding boxes. We minimize a standard detection objective
\begin{equation}
\min_{\phi}\ \mathcal{L}_{\mathrm{det}}\big(g_{\phi}(\mathbf{z}^{(L)}),\ \mathbf{y}\big),
\end{equation}
with focal classification loss and GIoU regression loss. This design keeps the number of trained parameters small, attributing gains primarily to the prompt-enhanced features.

\subsection{Why It Works: Prompts as Semantic Anchors}
\label{sec:why}

SPEM provides \emph{semantic anchors} by converting target-domain visual prototypes into prompts that guide a frozen backbone toward defect-relevant patterns at both shallow and mid-level layers. DAPA complements this by aligning prompt-conditioned features across domains, allowing knowledge captured by the source-trained detector to transfer more reliably. Together they promote fine-grained specialization and cross-domain consistency under a parameter-efficient training regime.

\section{Experiments}
\label{sec:experiments}

We conduct a comprehensive set of experiments to validate the effectiveness of our proposed framework. We first describe the experimental setup, including datasets, metrics, implementation details, and baseline categories. We then present main results, followed by ablation studies and qualitative analyses that provide deeper insight into robustness and behavior.

\subsection{Experimental Setup}

\paragraph{Datasets} We evaluate on four challenging, publicly available road defect benchmarks that exhibit substantial cross-domain variation: 
(1) \textbf{RDD}~\cite{arya2022rdd}, a multi-national dataset collected from Japan, India, and the Czech Republic under diverse conditions; 
(2) \textbf{CNRDD}~\cite{bianco2023pothole}, a high-resolution pothole dataset from Italy with significant class imbalance; 
(3) \textbf{TD-RD}~\cite{xiao2025tdrd}, featuring dense annotations and five damage categories across heterogeneous scenes; 
and (4) \textbf{CRDDC'22}~\cite{buslaev2022crddc}, a large-scale benchmark covering urban road environments.  
We adopt a strict \emph{single-source $\rightarrow$ single-target UDA protocol}: training is conducted on one labeled source domain (e.g., RDD-Japan) together with \emph{unlabeled} target images (e.g., TD-RD), and evaluation is performed on the target domain without using target labels. For few-shot analysis, we fine-tune the detector using a randomly sampled 5\% subset of labeled target data.

\begin{table*}[!ht]
\centering
\caption{Comparison with state-of-the-art methods under a unified protocol. All models use input resolution $512{\times}512$ and are evaluated on an NVIDIA A100 (FP16). FLOPs are computed at the same resolution; FPS is measured with batch size 1. \textbf{Pre (\%)} denotes class-averaged precision at IoU$\,=0.5$, respectively.}
\vspace{-4pt}
\label{tab:sota_final_exhaustive}
\resizebox{\textwidth}{!}{%
\begin{tabular}{lcccccccccccc}
\toprule
\multirow{2}{*}{\textbf{Model}} & \multicolumn{4}{c}{\textbf{TD-RD}} & \multicolumn{4}{c}{\textbf{CNRDD}} & \multicolumn{4}{c}{\textbf{CRDDC'22}} \\
\cmidrule(lr){2-5} \cmidrule(lr){6-9} \cmidrule(lr){10-13}
 & mAP (\%) ↑ & Pre (\%) ↑ & FLOPs (G) & FPS & mAP (\%) ↑ & Pre (\%) ↑ & FLOPs (G) & FPS & mAP (\%) ↑ & Pre (\%) ↑ & FLOPs (G) & FPS \\
\midrule
\multicolumn{13}{@{}l}{\textit{Category 1: Supervised Detectors (trained on source, evaluated on target)}} \\
\midrule
Faster R-CNN~\cite{ren2015faster}  & 74.6 & 76.6 & 94.3 & 10  & 20.3 & 28.1 & 94.3 & 10  & 39.9 & 45.3 & 94.3 & 10  \\
SSD~\cite{liu2016ssd}              & 76.6 & 71.1 & 60.9 & 14  & 18.5 & 27.4 & 60.9 & 14  & 38.7 & 44.2 & 60.9 & 14  \\
YOLOv5-s~\cite{yolov5}             & 85.6 & 84.6 & 15.8 & 111 & 22.5 & 30.7 & 15.8 & 111 & 42.1 & 46.4 & 15.8 & 111 \\
YOLOv6-s~\cite{li2022yolov6}       & 83.0 & 82.5 & 45.3 & 81  & 24.6 & 33.8 & 45.3 & 81  & 42.4 & 46.0 & 45.3 & 81  \\
YOLOv7~\cite{wang2023yolov7}       & 84.5 & 85.7 & 13.2 & 294 & 25.3 & 33.5 & 13.2 & 294 & 46.2 & 49.8 & 13.2 & 294 \\
YOLOv8-s~\cite{ultralytics2023yolov8}& 86.1 & 86.0 & 28.4 & 333 & 23.1 & 31.2 & 28.4 & 333 & 42.5 & 47.1 & 28.4 & 333 \\
YOLOv9-s~\cite{wang2024yolov9}     & 85.1 & 88.6 & 30.3 & 172 & 25.5 & 34.4 & 30.3 & 172 & 43.4 & 47.7 & 30.3 & 172 \\
YOLOv10-s~\cite{wang2024yolov10}   & 85.0 & 82.2 & 24.5 & 286 & 24.8 & 33.4 & 24.5 & 286 & 43.3 & 47.3 & 24.5 & 286 \\
YOLOS-s~\cite{fang2021yolos}       & 84.7 & 83.2 & 179  & 54  & 23.6 & 31.4 & 179  & 54  & 42.8 & 46.4 & 179  & 54  \\
RT-DETR~\cite{lv2023detrs}         & 87.7 & 87.7 & 60.0 & 159 & 24.2 & 32.5 & 60.0 & 159 & 43.1 & 48.2 & 60.0 & 159 \\
Lite-DERT~\cite{li2023lite}    & 86.1 & 85.2 & 151  & 75  & 24.3 & 33.0 & 151  & 75  & 42.3 & 46.9 & 151  & 75  \\
PP-PicoDet~\cite{yu2023pppicodet}  & 85.6 & 83.4 & 8.9  & 196 & 22.4 & 31.7 & 8.9  & 196 & 43.0 & 48.0 & 8.9  & 196 \\
MGD-YOLO~\cite{li2025mgdyolo}      & 87.0 & 86.1 & 34.2 & 175 & 25.0 & 34.1 & 34.2 & 175 & 43.6 & 48.5 & 34.2 & 175 \\
\midrule
\multicolumn{13}{@{}l}{\textit{Category 2: Self-Supervised Pre-training + Detection Head}} \\
\midrule
SimCLR~\cite{chen2020simclr}         & 86.5 & 85.4 & 32.1 & 195 & 26.3 & 34.8 & 32.1 & 195 & 45.2 & 50.1 & 32.1 & 195 \\
MoCo-v2~\cite{chen2020mocov2}        & 86.8 & 85.9 & 32.1 & 195 & 27.0 & 35.5 & 32.1 & 195 & 45.9 & 50.8 & 32.1 & 195 \\
SimSiam~\cite{chen2021simsiam}       & 87.1 & 86.2 & 32.0 & 196 & 27.5 & 36.1 & 32.0 & 196 & 46.2 & 51.1 & 32.0 & 196 \\
\midrule
\multicolumn{13}{@{}l}{\textit{Category 3: Cross-Domain Adaptation Methods + Detection Head}} \\
\midrule
DANN~\cite{ganin2016dann}           & 87.3 & 86.5 & 33.5 & 180 & 30.2 & 38.8 & 33.5 & 180 & 47.1 & 51.5 & 33.5 & 180 \\
MCD~\cite{saito2018mcd}             & 87.5 & 86.8 & 33.6 & 179 & 31.3 & 40.1 & 33.6 & 179 & 47.8 & 52.3 & 33.6 & 179 \\
CDTrans~\cite{xu2021cdtrans}        & 87.8 & 87.9 & 35.1 & 172 & 32.5 & 41.3 & 35.1 & 172 & 48.2 & 52.9 & 35.1 & 172 \\
\midrule
\multicolumn{13}{@{}l}{\textit{Category 4: Prompt Tuning Strategies (Supervised) + Detection Head}} \\
\midrule
VPT~\cite{jia2022vpt}               & 87.6 & 87.1 & 31.9 & 202 & 28.1 & 37.0 & 31.9 & 202 & 46.5 & 51.3 & 31.9 & 202 \\
MaPLe~\cite{khattak2023maple}       & 87.7 & 87.3 & 32.4 & 198 & 28.9 & 38.1 & 32.4 & 198 & 46.9 & 51.8 & 32.4 & 198 \\
\midrule
\textbf{\ours (Ours)}                   & \textbf{90.2} & \textbf{90.5} & \textbf{31.8} & \textbf{203} & \textbf{38.1} & \textbf{47.2} & \textbf{31.8} & \textbf{203} & \textbf{50.3} & \textbf{55.1} & \textbf{31.8} & \textbf{203} \\
\bottomrule
\end{tabular}
}
\end{table*}

\paragraph{Evaluation Metrics}
Following standard practice in detection and domain adaptation~\cite{li2025mgdyolo,xu2021cdtrans}, we report \textbf{mAP@50} and COCO-style \textbf{mAP@[.5:.95]}. We further analyze class-wise precision and recall to assess semantic discrimination. To ensure robustness, results are averaged over three runs with different seeds. Computational efficiency is reported in FLOPs and FPS, measured under a unified resolution ($512{\times}512$) on an NVIDIA A100 GPU with FP16 inference for all methods.

\vspace{-4pt}
\paragraph{Implementation Details}
We use a ViT-B/16 backbone~\cite{dosovitskiy2020vit} pre-trained on ImageNet-1K, kept frozen during self-supervised training. SPEM uses $K=10$ prototypes obtained by K-means clustering on PCA-reduced target patch embeddings, projected back to the ViT dimension $D=768$ via a two-layer MLP with GELU. Prompts are injected at both the input layer and the 6th transformer layer. We optimize with AdamW (lr=$3\times10^{-4}$, weight decay=$1\times10^{-4}$) and a cosine schedule for 200 epochs, batch size 64. Downstream detection uses a lightweight three-layer convolutional head trained for 50 epochs with 500 labeled source images. Unless stated otherwise, all hyperparameters follow this default setup.

\vspace{-4pt}
\paragraph{Baselines}
We compare \ours against four categories of strong baselines: \textit{Supervised Detectors}: Faster R-CNN~\cite{ren2015faster}, SSD~\cite{liu2016ssd}, the YOLO family (v5~\cite{yolov5}, v6~\cite{li2022yolov6}, v7~\cite{wang2023yolov7}, v8~\cite{ultralytics2023yolov8}, v9~\cite{wang2024yolov9}, v10~\cite{wang2024yolov10}), Transformer-based detectors (YOLOS~\cite{fang2021yolos}, RT-DETR~\cite{lv2023detrs}, Lite-DERT~\cite{li2023lite}), PP-PicoDet~\cite{yu2023pppicodet}, and MGD-YOLO~\cite{li2025mgdyolo}. \textit{Self-Supervised Methods}: SimCLR~\cite{chen2020simclr}, MoCo-v2~\cite{chen2020mocov2}, SimSiam~\cite{chen2021simsiam}. \textit{Cross-Domain Adaptation Methods}: source-only training, DANN~\cite{ganin2016dann}, MCD~\cite{saito2018mcd}, and CDTrans~\cite{xu2021cdtrans}. \textit{Prompt Tuning Strategies}: VPT~\cite{jia2022vpt} and MaPLe~\cite{khattak2023maple}.

\subsection{Comparison with State-of-the-Art Methods}

Table~\ref{tab:sota_final_exhaustive} compares \ours with strong supervised detectors, self-supervised pre-training, cross-domain adaptation~(CDA), and prompt-tuning strategies under a unified evaluation protocol. \ours attains the highest mAP on the three target datasets, while maintaining a competitive efficiency profile. We summarize the key observations below.

\vspace{-8pt}
\paragraph{Supervised detectors reveal a pronounced domain gap}
Modern supervised detectors (e.g., RT-DETR~\cite{lv2023detrs}) achieve high accuracy on source-like data (e.g., 87.7\% mAP on TD-RD), yet their performance drops considerably on cross-domain targets (24.2\% on CNRDD; 43.1\% on CRDDC'22). This aligns with prior evidence that \emph{without explicit adaptation}, in-domain gains do not translate to strong out-of-domain generalization~\cite{wang2022generalizing,zhou2022domain}.

\vspace{-4pt}
\paragraph{Adaptation is necessary but alignment granularity matters}
SSL baselines (SimCLR/MoCo/SimSiam) improve over purely supervised training on target domains (e.g., SimSiam 27.5\% vs.\ YOLOv5-s 22.5\% on CNRDD), indicating the value of unlabeled data. CDA methods that \emph{explicitly} reduce distribution shifts bring larger gains; CDTrans~\cite{xu2021cdtrans} is a strong reference (32.5\% on CNRDD; 48.2\% on CRDDC'22). However, these approaches typically align \emph{global} statistics, which may downweight sparse, fine-grained defect cues critical to road inspection.

\vspace{-4pt}
\paragraph{\ours advances accuracy with prompt-enhanced alignment}
\ours achieves 90.2\% on TD-RD, 38.1\% on CNRDD, and 50.3\% on CRDDC'22. The gains over the strongest CDA baseline (CDTrans) are +5.6 mAP on CNRDD and +2.1 mAP on CRDDC'22. We attribute these improvements to the synergy between \textbf{SPEM}, which turns unlabeled \emph{target} imagery into task-aware prompts to emphasize defect-relevant patterns, and \textbf{DAPA}, which aligns \emph{prompt-conditioned} representations. This targeted alignment is consistent with observations that class-/semantics-aware objectives can be more effective than global alignment for challenging transfers~\cite{tzeng2017adversarial,saito2018mcd}.

\subsection{Ablation Studies}

We conduct ablations to quantify the contribution of each component, examine key prompt design choices, and assess sensitivity to loss weights. Unless otherwise noted, results are averaged over three runs with different seeds; all numbers are mAP (\%) on target domains.

\begin{table}[t]
\centering
\small
\caption{Ablations on (a) core components, (b) prompt design, and (c) loss weights. All numbers are mAP (\%) on target domains, averaged over three seeds. In (a), the first row is a \emph{source-only} lower bound (no SSL, no SPEM, no DAPA). The selected variants are \hl{highlighted in blue}.}
\vspace{-4pt}
\label{tab:ablation_all}
\renewcommand{\arraystretch}{0.85}
\begin{subtable}{\columnwidth}\centering
\caption{Core components.}
\label{tab:ablation_main}
\resizebox{0.9\textwidth}{!}{%
\begin{tabular}{ccc cc}
\toprule
\multicolumn{3}{c}{\textbf{Components}} & \multicolumn{2}{c}{\textbf{mAP (\%) $\uparrow$}} \\
\cmidrule(lr){1-3} \cmidrule(lr){4-5}
SSL & SPEM & DAPA & CNRDD & CRDDC'22 \\
\midrule
 &  &  & 23.1 & 42.5 \\
\cmark &  &  & 27.5 & 46.2 \\
\cmark & \cmark &  & 33.8 & 48.1 \\
\cmark &  & \cmark & 34.5 & 48.7 \\
\rowcolor{wacvblue!10} \cmark & \cmark & \cmark & \textbf{38.1} & \textbf{50.3} \\
\bottomrule
\end{tabular}
}
\end{subtable}

\vspace{0.5em}

\begin{subtable}{\columnwidth}\centering
\caption{Prompt design.}
\label{tab:ablation_design}
\resizebox{0.9\textwidth}{!}{%
\begin{tabular}{lcc}
\toprule
\textbf{Parameter} & \textbf{CNRDD (\%)} $\uparrow$ & \textbf{CRDDC'22 (\%)} $\uparrow$ \\
\midrule
\multicolumn{3}{@{}l}{\textit{Number of prompts ($K$)}} \\
$K=1$     & 31.5 & 47.3 \\
$K=5$     & 36.8 & 49.5 \\
\rowcolor{wacvblue!10} \textbf{$K=10$} & \textbf{38.1} & \textbf{50.3} \\
$K=15$    & 37.7 & 50.1 \\
\midrule
\multicolumn{3}{@{}l}{\textit{Injection depth}} \\
Shallow (L0)     & 36.5 & 49.2 \\
Mid (L6)         & 37.1 & 49.6 \\
\rowcolor{wacvblue!10} Shallow+Mid & \textbf{38.1} & \textbf{50.3} \\
\bottomrule
\end{tabular}
}
\end{subtable}

\vspace{0.5em}

\begin{subtable}{\columnwidth}\centering
\caption{Loss weights.}
\label{tab:ablation_lambda}
\resizebox{\textwidth}{!}{%
\begin{tabular}{llcc}
\toprule
$\lambda_1$ (SPEM) & $\lambda_2$ (DAPA) & \textbf{CNRDD (\%)} $\uparrow$ & \textbf{CRDDC'22 (\%)} $\uparrow$ \\
\midrule
0.1 & 0.5 & 36.2 & 49.1 \\
\rowcolor{wacvblue!10}  1.0 & 0.5 & \textbf{38.1} & \textbf{50.3} \\
2.0 & 0.5 & 37.8 & 50.1 \\
\midrule
1.0 & 0.1 & 36.9 & 49.4 \\
\rowcolor{wacvblue!10} 1.0 & 0.5 & \textbf{38.1} & \textbf{50.3} \\
1.0 & 1.0 & 37.4 & 49.9 \\
\bottomrule
\end{tabular}
}
\end{subtable}

\end{table}

\paragraph{Core components}
Table~\ref{tab:ablation_main} isolates each module's contribution. Relative to the \emph{source-only} lower bound (23.1/42.5), adding SSL yields +4.4/+3.7 mAP on CNRDD/CRDDC'22, confirming the value of unlabeled data. Building on SSL, \textbf{SPEM} improves another +6.3/+1.9 by converting target-domain structure into task-aware prompts; \textbf{DAPA} alone contributes +7.0/+2.5 via cross-domain alignment. Combining both delivers the largest gains (38.1/50.3), indicating complementary effects: SPEM refines \emph{what} to emphasize, while DAPA makes these features \emph{transferable}.

\vspace{-4pt}
\paragraph{Prompt design}
As shown in Table~\ref{tab:ablation_design}, increasing the number of prompts from $K{=}1$ to $K{=}10$ steadily boosts performance (e.g., +6.6 on CNRDD), suggesting higher expressive capacity helps capture diverse defect patterns. Beyond $K{=}10$, the curve plateaus slightly—consistent with prompt tuning literature where excessive parameters may model spurious correlations~\cite{jia2022vpt, anonymous2025cadvae, ma2025stochastic, chenrui2025Learn, ma2025beyond}. For injection depth, a \textbf{Shallow+Mid} strategy outperforms single-point injection, aligning with evidence that ViT layers capture progressively abstract features~\cite{dascoli2021vit_learn}; early prompts steer low-level textures, mid-level prompts consolidate defect semantics.

\vspace{-4pt}
\paragraph{Sensitivity to loss weights}
Table~\ref{tab:ablation_lambda} indicates that \ours is robust to reasonable variations of $(\lambda_1,\lambda_2)$. Performance peaks near $(1.0,0.5)$, but deviating to $(0.5{\sim}2.0,\,0.1{\sim}1.0)$ results in $\leq\!0.6$ mAP change on either target, suggesting the method does not hinge on fragile hyperparameter tuning—an attractive property for practical deployment.

\vspace{-4pt}
\subsection{Analysis of Prompt Design}
The design of the self-supervised prompts is central to \ours's effectiveness. We therefore conduct targeted ablations on the number of prompt tokens ($K$) and the depth at which they are injected into the ViT backbone. All results are averaged over three runs to ensure stability.

\begin{figure}[t!]
\centering
\includegraphics[width=0.95\columnwidth]{ 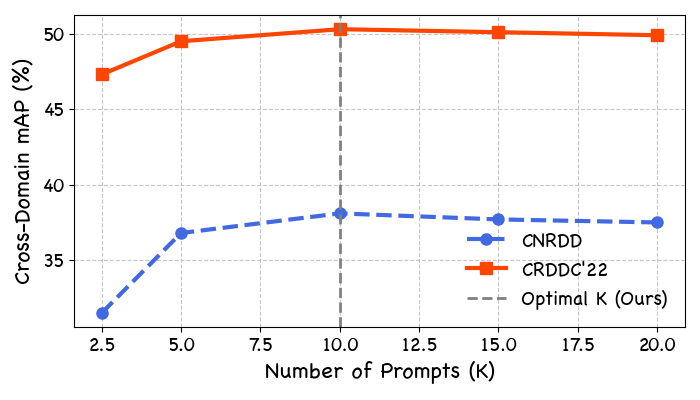}
\vspace{-4pt}
\caption{Impact of the number of prompts ($K$) on cross-domain mAP (\%). Performance peaks at $K=10$, after which it slightly declines. Results are averaged over three runs.}
\label{fig:prompt_k}
\vspace{-4pt}
\end{figure}

\vspace{-4pt}
\paragraph{How many prompts?}
Figure~\ref{fig:prompt_k} shows that too few prompts ($K{<}10$) lack the capacity to capture the diverse visual patterns of defects, resulting in lower accuracy. Increasing $K$ improves mAP steadily, with the best trade-off reached at $K{=}10$. Beyond this point, adding more prompts yields marginal or negative gains. This plateau effect is consistent with prior prompt tuning literature~\cite{jia2022vpt}, where excessive parameters may overfit to spurious correlations or noise. We therefore adopt $K{=}10$ in all main experiments.

\begin{figure}[t!]
\centering
\includegraphics[width=0.48\textwidth]{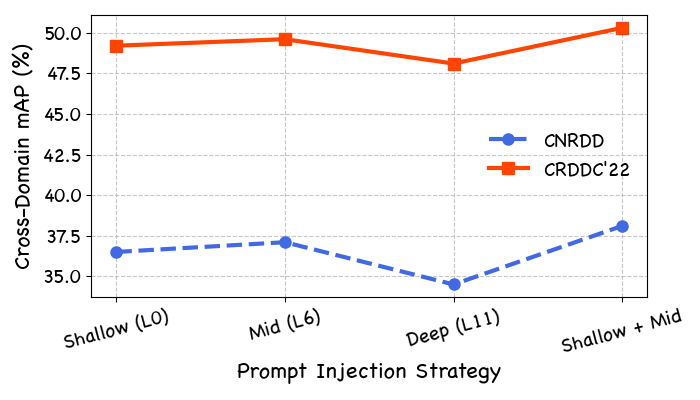}
\vspace{-8pt}
\caption{Impact of injection depth on cross-domain mAP (\%). A multi-stage strategy (\emph{Shallow+Mid}) consistently outperforms single-layer injection.}
\label{fig:prompt_depth}
\vspace{-4pt}
\end{figure}

\vspace{-4pt}
\paragraph{Where to inject?}
As shown in Figure~\ref{fig:prompt_depth}, injection depth plays a crucial role. A shallow-only strategy (input layer) already provides meaningful guidance, but combining shallow and mid-level layers yields the strongest performance. This suggests that prompts benefit from influencing the feature hierarchy at multiple levels: early prompts bias low-level textures and edges, while mid-level prompts help assemble these into more complex semantics. This observation is consistent with studies showing that ViT layers capture progressively more abstract features~\cite{dascoli2021vit_learn}. We thus adopt a \emph{Shallow+Mid} strategy by default.

\subsection{Visualization Analysis}
\begin{figure}[!th]
  \centering
  \includegraphics[width=0.9\linewidth]{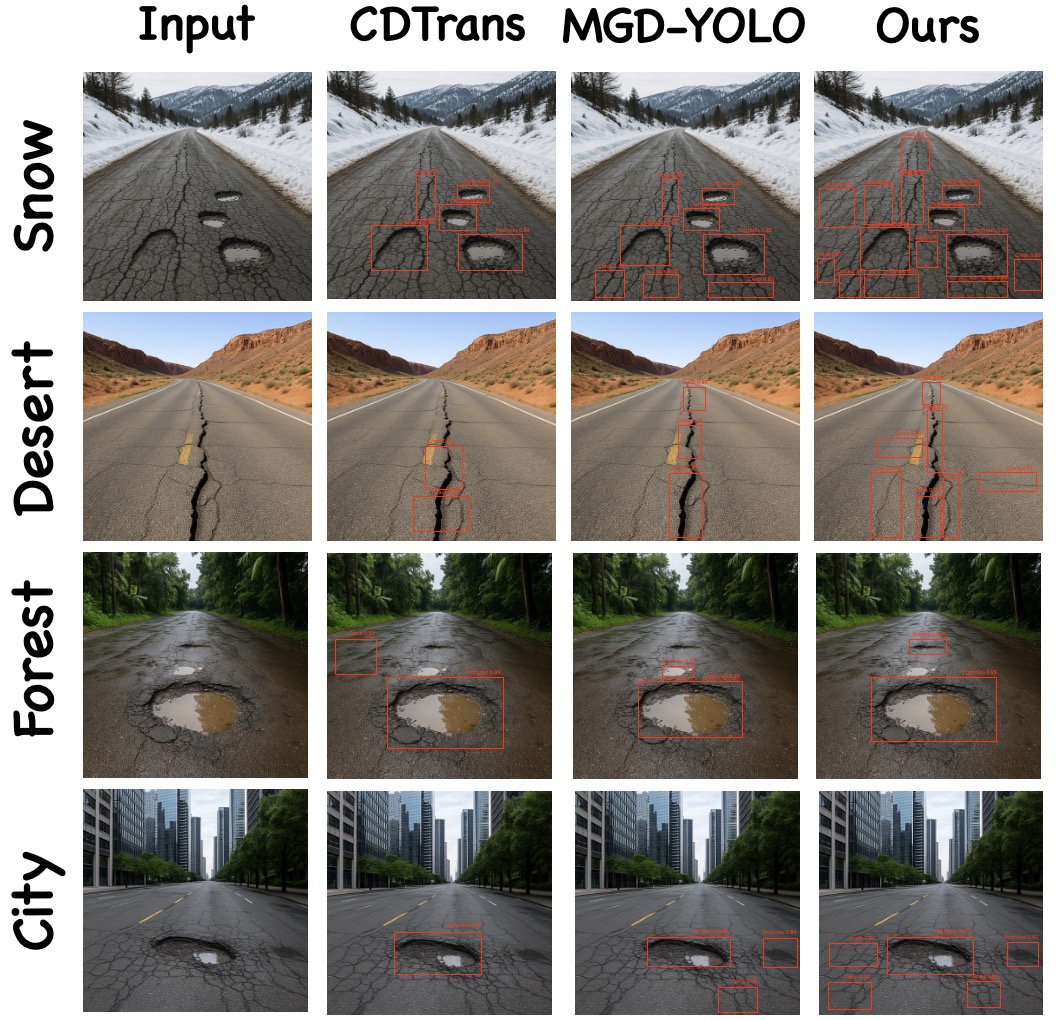}
  \vspace{-2mm}
  \caption{Detection results of different state-of-the-art methods across multiple domains (Snow, Desert, Forest, and City). Our method achieves more robust detection under diverse environments compared with CDTrans and MGD-YOLO.}
  \label{fig:domain_results}
  \vspace{-3mm}
\end{figure}

\begin{figure}[!th]
  \centering
  \includegraphics[width=0.9\linewidth]{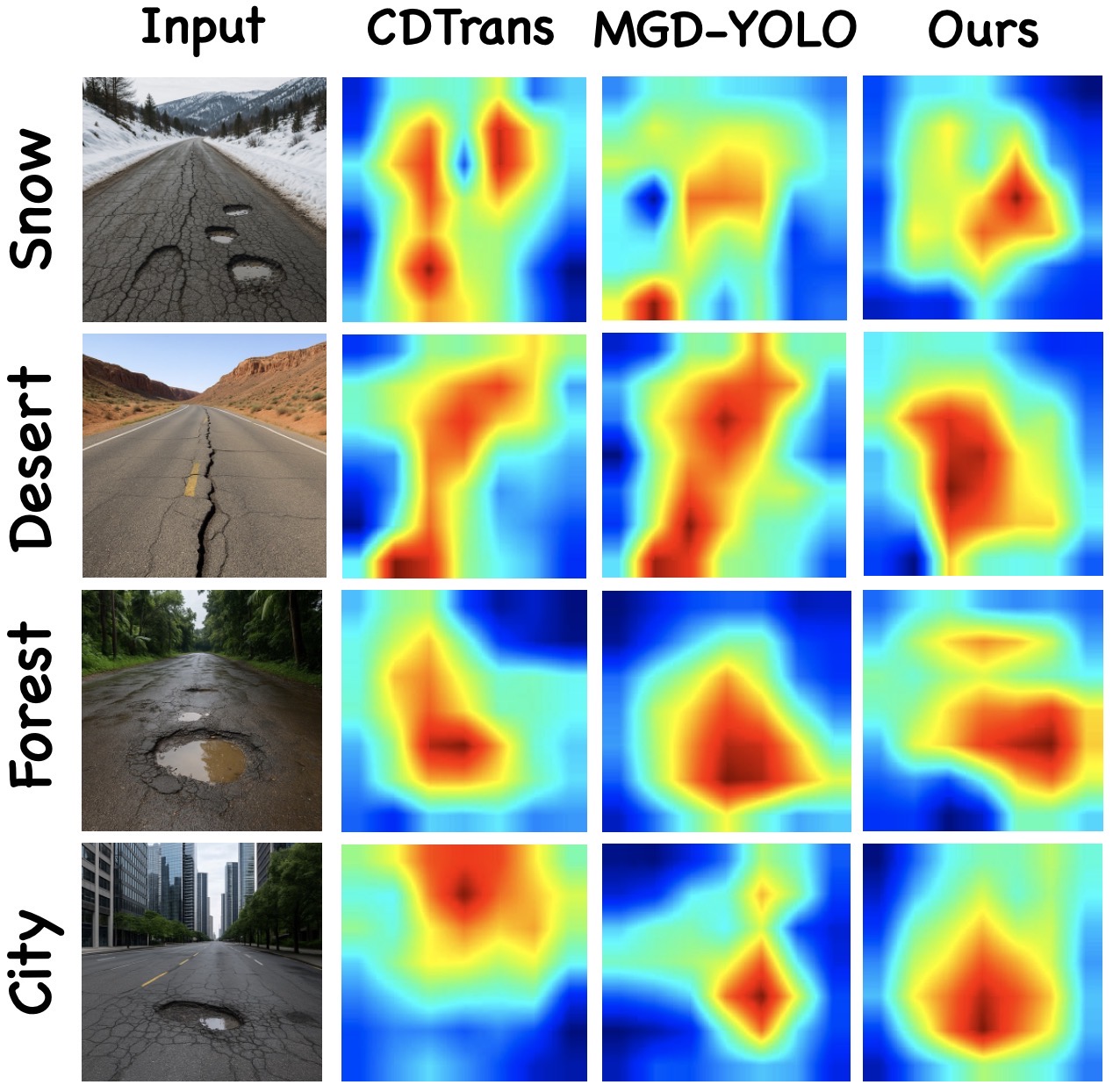}
  \vspace{-2mm}
  \caption{\textbf{Heatmap++ visualization of focus regions across domains.}
  We visualize the core focus regions of different methods under four domains (Snow, Desert, Forest, City).
  Heatmap++ highlights where models attend when predicting defects.
  Compared to CDTrans and MGD-YOLO, our method concentrates more precisely on defect areas (cracks/potholes) and suppresses background textures, showing superior cross-domain localization.}
  \label{fig:heatmap_focus}
  \vspace{-3mm}
\end{figure}

To better understand how different models behave in cross-domain cases, we show two kinds of visualizations in Fig.~\ref{fig:domain_results} and Fig.~\ref{fig:heatmap_focus}. 

\vspace{-4pt}
\paragraph{Detection Visualization}
Fig.~\ref{fig:domain_results} compares the detection outputs. CDTrans produces bounding boxes with low precision and often misses small cracks. MGD-YOLO covers more areas but still draws redundant or imprecise boxes. Our model gives more stable outputs, with tighter and more accurate boxes around the real defects on all domains.

\vspace{-4pt}
\paragraph{Heatmap++ Visualization}
Fig.~\ref{fig:heatmap_focus} gives the focus maps of CDTrans, MGD-YOLO, and our model on four domains (Snow, Desert, Forest, City). CDTrans and MGD-YOLO often spread attention over wide background areas or non-defect parts, which makes the focus unstable. Our model instead shows compact maps centered on cracks and potholes, while ignoring lane markings or shadows. This shows that the prompts in our method guide the model to focus more on defect-related regions.

\subsection{Few-Shot Fine-Tuning Performance}

Although \ours is primarily designed for zero-shot cross-domain adaptation, we further evaluate its \emph{data efficiency} in few-shot scenarios. This setting reflects practical deployments, where a small portion of target-domain labels may be available to boost performance at low annotation cost. We fine-tune only the detection heads of our model and selected baselines on randomly sampled $1\%$, $5\%$, and $10\%$ subsets of the labeled CRDDC'22 training data, while keeping all backbones frozen to isolate the quality of the learned representations.

\begin{table}[!th]
\centering
\caption{Few-shot fine-tuning performance on CRDDC'22. mAP (\%) is reported as a function of the percentage of labeled target data. Zero-shot corresponds to $0\%$ labeled target data. Results show averages over three runs.}
\vspace{-8pt}
\label{tab:few_shot}
\resizebox{0.45\textwidth}{!}{%
\begin{tabular}{lcccc}
\toprule
\multirow{2}{*}{\textbf{Model}} & \multicolumn{4}{c}{\textbf{mAP (\%)} $\uparrow$} \\
\cmidrule(lr){2-5}
& 0\% & 1\% & 5\% & 10\% \\
\midrule
Supervised (YOLOv8-s) & 42.5 & 45.0 & 48.5 & 50.5 \\
SSL (SimSiam)         & 46.2 & 49.5 & 52.0 & 53.8 \\
CDA (CDTrans)         & 48.2 & 51.0 & 53.5 & 55.0 \\
\textbf{\ours (Ours)} & \textbf{50.3} & \textbf{54.5} & \textbf{57.0} & \textbf{58.5} \\
\bottomrule
\end{tabular}
\vspace{-18pt}
}
\end{table}

\vspace{-4pt}
\paragraph{Results and discussion}
Table~\ref{tab:few_shot} highlights the data-efficiency of \ours. At the zero-shot level, our model already outperforms all baselines, confirming the strength of its self-supervised pre-training. As target labels are introduced, the gap widens: with only \textbf{1\%} labeled target data, \ours reaches 54.5\% mAP, while CDTrans requires \textbf{5\%} labels to approach a comparable score. This corresponds to a roughly \textbf{5$\times$ gain in label efficiency}. Even with 10\% labels, \ours maintains a consistent lead. These results indicate that the proposed self-supervised prompting not only enhances robustness to domain shift but also provides an excellent initialization for rapid and low-cost adaptation in practical settings.

\subsection{Evaluation of Cross-Domain Robustness}

A robust adaptation framework should generalize not only to a few held-out benchmarks but also under diverse training conditions and input perturbations. We therefore evaluate \ours in two complementary settings: varying the source domain and applying synthetic corruptions.

\begin{table}[!t]
\centering
\caption{Multi-source $\rightarrow$ multi-target evaluation. \ours exhibits more stable performance than CDTrans when varying the source domain. C'22 denotes CRDDC'22.}
\vspace{-8pt}
\label{tab:robustness_source}
\resizebox{0.48\textwidth}{!}{%
\begin{tabular}{llccccc}
\toprule
\multirow{2}{*}{\textbf{Method}} & \multirow{2}{*}{\textbf{Source}} & \multicolumn{5}{c}{\textbf{Target Domain mAP (\%)} $\uparrow$} \\
\cmidrule(lr){3-7}
&& Japan & India & Czech & CNRDD & C'22 \\
\midrule
\multirow{3}{*}{CDTrans} & RDD-Japan & ---  & 86.2 & 85.0 & 32.5 & 48.2 \\
& RDD-India & 84.8 & ---  & 84.1 & 29.8 & 46.5 \\
& RDD-Czech & 84.5 & 83.9 & ---  & 29.1 & 46.1 \\
\midrule
\multirow{3}{*}{\textbf{\ours (Ours)}} & RDD-Japan & ---  & \textbf{89.5} & \textbf{88.7} & \textbf{38.1} & \textbf{50.3} \\
& RDD-India & \textbf{88.9} & ---  & \textbf{88.2} & \textbf{37.4} & \textbf{49.8} \\
& RDD-Czech & \textbf{88.5} & \textbf{87.9} & ---  & \textbf{37.1} & \textbf{49.5} \\
\bottomrule
\end{tabular}
}
\vspace{-8pt}
\end{table}

\paragraph{Robustness to source variation}
To test whether our method's effectiveness depends on a specific source, we conduct a multi-source, multi-target evaluation. We train \ours and the strongest CDA baseline (CDTrans) on three different RDD sources (Japan, India, Czech) and evaluate them on all other unseen targets. Results in Table~\ref{tab:robustness_source} show that \ours maintains consistently high performance across all pairs. For example, when trained on RDD-India or RDD-Czech, the mAP on CNRDD and CRDDC'22 remains within 1 point of the RDD-Japan source model. In contrast, CDTrans shows larger fluctuations, with up to a 3.4-point drop on CNRDD. These findings indicate that self-supervised prompt generation provides stable generalization regardless of the chosen source domain.

\begin{figure}[h!]
    \centering
    \includegraphics[width=\columnwidth]{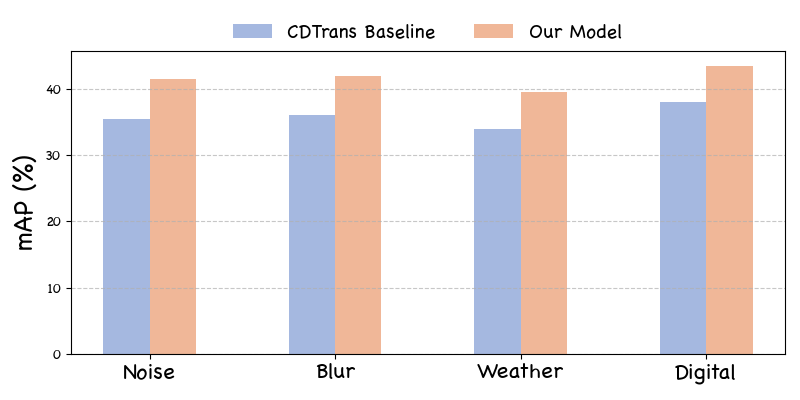}
    \caption{Performance on CRDDC'22 under four corruption categories. \ours consistently suffers smaller drops than CDTrans, indicating stronger robustness.}
    \label{fig:robustness_corruption}
\end{figure}

\paragraph{Robustness to common corruptions}
We further stress-test representation stability under common perturbations following the ImageNet-C protocol. Four categories of corruptions (noise, blur, weather, digital artifacts) are applied to CRDDC'22 test images. Figure~\ref{fig:robustness_corruption} compares mAP drops of \ours and CDTrans. Both methods degrade, but \ours consistently shows smaller losses. For instance, under \emph{Weather} corruptions, \ours drops by 10.8 points versus 14.2 for CDTrans. This suggests that prompt-enhanced alignment yields features that rely less on superficial textures, enhancing resilience to realistic degradations.

\section{Conclusion}

We presented \ours, a self-supervised framework for cross-domain road damage detection. The key idea is to generate defect-aware prompts from unlabeled target images (SPEM) and align the resulting prompt-conditioned features across domains (DAPA). This combination provides both task-specific specialization and cross-domain consistency within a parameter-efficient design. Through extensive experiments, \ours consistently outperforms supervised, SSL, CDA, and prompt-tuning baselines. It demonstrates strong zero-shot generalization, stable robustness across diverse conditions, and superior data efficiency in few-shot adaptation. Beyond road damage detection, our study suggests that \emph{self-supervised prompting} is a promising direction for building adaptive inspection systems in safety-critical domains. Future extensions include exploring source-free adaptation, multimodal prompt integration, and applications to broader visual inspection tasks where robustness and scalability are crucial.

\section*{Acknowledgment}

This manuscript was co-authored by Oak Ridge National Laboratory (ORNL), operated by UT-Battelle, LLC under Contract No. DE-AC05-00OR22725 with the U.S. Department of Energy. Any subjective views or opinions expressed in this paper do not necessarily represent those of the U.S. Department of Energy or the United States Government.

{
    \small
    \bibliographystyle{ieeenat_fullname}
    \bibliography{main}
}

\maketitlesupplementary

\subsection*{A. Prompt Generation and Clustering Strategy}
\label{supp:spem_details}

We detail the unsupervised procedure for deriving domain-adaptive visual prompts from unlabeled target data $X^t$. The pipeline consists of four steps: patch feature extraction, dimensionality reduction, prototype discovery, and prompt projection. All steps use fixed random seeds and are repeated with three seeds for stability.

\paragraph{Step A1: Patch-level feature extraction (frozen ViT).}
For each $\mathbf{x}_i^t\!\in\!X^t$, a frozen ViT-B/16 encoder (embedding dimension $D{=}768$) produces patch tokens
$\mathbf{z}_i^{(0)}\!\in\!\mathbb{R}^{N\times D}$, with $N{=}196$ (for $224{\times}224$ input). We discard class tokens and retain only patch tokens.

\paragraph{Step A2: Dimensionality reduction (global over $X^t$).}
We aggregate \emph{all} target patch tokens $\{\mathbf{z}^{(0)}_{i,n}\}$ and apply PCA to obtain
$\tilde{\mathbf{z}}^{(0)}_{i,n}\!\in\!\mathbb{R}^{d'}$ with $d'{=}50$. PCA is fit once on $X^t$ (whiten=False). This global scheme avoids batch-level drift.

\paragraph{Step A3: Visual prototype discovery (K-means over $X^t$).}
We run K-means on $\{\tilde{\mathbf{z}}^{(0)}_{i,n}\}$ to obtain $K$ centroids
$\mathcal{C}{=}\{\mathbf{c}_k\}_{k=1}^K$, $\mathbf{c}_k\!\in\!\mathbb{R}^{d'}$.
We use k-means++ initialization, max\_iter=300, n\_init=10. Following our ablations, $K{=}10$ offers the best capacity/efficiency trade-off. In practice, centroids are computed \emph{once} per target domain; optional re-computation every 50 epochs gave negligible changes ($<0.2$ mAP).

\paragraph{Step A4: Learnable prompt projection and injection.}
A two-layer MLP with GELU maps prototypes back to the ViT space:
\[
\mathbf{P}^t \;=\; \mathrm{MLP}_{\theta_p}(\mathcal{C}) \in \mathbb{R}^{K\times D},\quad D{=}768.
\]
At injection layers (Shallow $L{=}0$ and Mid $L{=}6$), we augment token sequences by concatenation
\[
\tilde{\mathbf{z}}^{(l-1)} \;=\; [\mathbf{P}^t;\ \mathbf{z}^{(l-1)}]\in\mathbb{R}^{(K+N)\times D}.
\]
Only $\theta_p$ is trainable in SPEM; the ViT remains frozen. We use the contrastive prompt loss from the main paper to encourage semantic consistency of prompts.

\paragraph{Reproducibility notes.}
All clustering uses \texttt{faiss}/\texttt{scikit-learn} with fixed seeds $\{0,1,2\}$. Unless stated, we report the mean over three runs. We did not observe sensitivity to k-means++ vs.\ random init beyond $\pm0.1$ mAP.


\subsection*{B. Theoretical Justification for DAPA}
\label{supp:dapa_theory}

\subsubsection*{B.1. Domain adaptation bound}
Let $D_s,D_t$ be source/target distributions and $h\!\in\!\mathcal{H}$. The expected risks are
$\epsilon_s(h)\!=\!\mathbb{E}_{(x,y)\sim D_s}[\mathbb{1}(h(x)\!\neq\!y)]$ and
$\epsilon_t(h)\!=\!\mathbb{E}_{(x,y)\sim D_t}[\mathbb{1}(h(x)\!\neq\!y)]$.
The target risk is bounded~\cite{ben2010theory} by
\begin{equation}
\label{eq:da_bound_supp}
\epsilon_t(h) \;\le\; \hat{\epsilon}_s(h)\;+\; d_{\mathcal{H}\Delta\mathcal{H}}(D_s,D_t)\;+\;\lambda,
\vspace{-2pt}
\end{equation}
where $d_{\mathcal{H}\Delta\mathcal{H}}$ measures domain discrepancy and $\lambda$ is the error of the ideal joint hypothesis. Hence, reducing the divergence term is crucial for target generalization.

\subsubsection*{B.2. MMD as a tractable discrepancy}
Given samples $X^s,X^t$, the squared MMD in an RKHS $\mathcal{H}_k$ is
\begin{equation}
\mathrm{MMD}^2(D_s,D_t)=\left\| \mathbb{E}_{x\sim D_s}[\phi(x)]-\mathbb{E}_{y\sim D_t}[\phi(y)] \right\|^2_{\mathcal{H}_k},
\end{equation}
with empirical estimate in Eq.~(13) of the main paper.

\subsubsection*{B.3. DAPA as linear-kernel MMD (prompt-enhanced space)}
For the linear kernel $k(x,y){=}x^\top y$, $\phi$ is identity and
\[
\widehat{\mathrm{MMD}}^2_{\text{lin}}(X^s,X^t)=\big\|\hat{\mathbb{E}}[X^s]-\hat{\mathbb{E}}[X^t]\big\|_2^2.
\]
Applying to prompt-enhanced representations $\mathbf{h}^s,\mathbf{h}^t$ with a projection head $f_p(\cdot)$ yields our DAPA loss:
\begin{equation}
\label{eq:dapa_supp}
\mathcal{L}_{\mathrm{DAPA}}=\left\|\mathbb{E}_{\mathbf{x}^s}[f_p(\mathbf{h}^s)]-\mathbb{E}_{\mathbf{x}^t}[f_p(\mathbf{h}^t)]\right\|_2^2,
\end{equation}
which directly minimizes a linear-kernel MMD in the \emph{prompt-conditioned} space.

\paragraph{Optional extensions.}
RBF-MMD with multi-bandwidth kernels, CORAL, HSIC, or class-conditional MMD (using prototype-induced pseudo-classes) are drop-in replacements; we found linear MMD most efficient and sufficiently effective in practice.


\subsection*{C. Training Objective, Algorithm, and Hyperparameters}
\label{supp:train_opt}

\subsubsection*{C.1. Composite loss}
We optimize
\begin{equation}
\label{eq:l_total_supp}
\mathcal{L}_{\text{total}}=\mathcal{L}_{\text{ssl}}+\lambda_1\mathcal{L}_{\text{prompt}}+\lambda_2\mathcal{L}_{\text{DAPA}},
\end{equation}
where $\mathcal{L}_{\text{ssl}}$ follows SimSiam~\cite{chen2021simsiam}; $\mathcal{L}_{\text{prompt}}$ is the InfoNCE-style prompt consistency; $\mathcal{L}_{\text{DAPA}}$ is Eq.~\eqref{eq:dapa_supp}.

\subsubsection*{C.2. Self-supervised pre-training loop (pseudocode)}

\begin{algorithm}[h]
\caption{Self-supervised pre-training with SPEM \& DAPA (frozen ViT)}
\label{alg:pretrain}
\begin{algorithmic}[1]
\State \textbf{Input:} unlabeled $X^s,X^t$; frozen ViT $f$; projector $g$, predictor $h$; prompt MLP $\theta_p$; projection head $f_p$
\State \textbf{Init:} PCA on $X^t$; K-means on PCA features $\Rightarrow \mathcal{C}$; $\mathbf{P}^t=\mathrm{MLP}_{\theta_p}(\mathcal{C})$
\For{epoch $=1\dots T$}
  \State Sample mini-batches $(\mathcal{B}_s,\mathcal{B}_t)$
  \State Inject $\mathbf{P}^t$ at layers $L{=}0,6$; obtain prompt-enhanced features $\mathbf{h}^s,\mathbf{h}^t$
  \State Compute $\mathcal{L}_{\text{ssl}}$ on $(\mathcal{B}_s\cup \mathcal{B}_t)$, $\mathcal{L}_{\text{prompt}}$ on $\mathcal{B}_t$, $\mathcal{L}_{\text{DAPA}}$ via Eq.~\eqref{eq:dapa_supp}
  \State Update $\theta_p$, $g$, $h$, $f_p$ by AdamW; keep $f$ frozen
  \If{epoch \% 50 == 0} \State (optional) recompute $\mathcal{C}$ on $X^t$ and refresh $\mathbf{P}^t$ \EndIf
\EndFor
\end{algorithmic}
\end{algorithm}

\subsubsection*{C.3. Trainable parameters and memory}
Trainable modules: prompt MLP ($\theta_p$), SSL projector/predictor ($g,h$), DAPA head ($f_p$). ViT is frozen throughout pre-training and downstream fine-tuning. On an A100 80GB, batch size 64 fits comfortably with FP16; peak memory $\approx$ 18–22GB for $224{\times}224$ inputs.

\subsection*{D. Downstream Head: Architecture and Fine-tuning}
\label{supp:head}

\paragraph{Design rationale.}
We adopt a minimalist head to attribute gains to \emph{prompt-enhanced} features rather than high-capacity decoders. The ViT backbone is frozen, and only a small three-stage convolutional head is trained, which keeps trainable parameters and memory footprint low while preserving fair attribution to the learned representations.

\paragraph{Input feature processing.}
Given the frozen ViT-B/16 outputs $\mathbf{z}^{(L)} \in \mathbb{R}^{N \times D}$ with $N{=}196$ tokens (for $224{\times}224$ input, $16{\times}16$ patch) and $D{=}768$, we discard the \texttt{[CLS]} and any prompt tokens, reshape patch tokens to a feature map $\mathbf{F}\in\mathbb{R}^{14\times 14\times 768}$, and feed it to the detection head $g_{\phi}(\cdot)$.

\paragraph{Architecture.}
The head consists of two light convolutional blocks followed by a $1{\times}1$ prediction layer producing $(C{+}4)$ channels per spatial location, where $C$ is the number of defect classes and $4$ are bounding-box parameters (e.g., $(\Delta x,\Delta y,\Delta w,\Delta h)$ in our implementation). BatchNorm (BN) and GELU are used as noted. The layer-by-layer specification is provided in Table~\ref{tab:head_arch}. We deliberately avoid multi-scale pyramids/decoders to keep the design minimal.

\begin{table*}[h!]
\centering
\caption{Layer-by-layer specification of the lightweight detection head $g_{\phi}(\cdot)$. $C$ denotes the number of classes. Parameter counts exclude BN affine terms for brevity.}
\label{tab:head_arch}
\resizebox{\textwidth}{!}{%
\begin{tabular}{l @{\hskip 72pt} l @{\hskip 64pt} l}
\toprule
\textbf{Layer} & \textbf{Operator (stride, pad)} & \textbf{Output shape} \\
\midrule
Input Feature Map & --- & $14 \times 14 \times 768$ \\
Conv Block 1 & Conv $3{\times}3$, s=1, p=1 $\rightarrow$ 384 \ + BN + GELU & $14 \times 14 \times 384$ \\
Conv Block 2 & Conv $1{\times}1$, s=1, p=0 $\rightarrow$ 128 \ + GELU & $14 \times 14 \times 128$ \\
Prediction Head & Conv $1{\times}1$, s=1, p=0 $\rightarrow (C{+}4)$ & $14 \times 14 \times (C{+}4)$ \\
\midrule
\multicolumn{3}{@{}l}{\emph{Parameter counts (approx.):}} \\
\multicolumn{3}{@{}l}{Conv1: $3{\times}3{\times}768{\times}384 \approx 2.65$M;\quad Conv2: $1{\times}1{\times}384{\times}128 \approx 49$K;} \\
\multicolumn{3}{@{}l}{Pred: $1{\times}1{\times}128{\times}(C{+}4) \approx 128(C{+}4)$.\quad Total $\approx 2.70$M $+$ BN.} \\
\bottomrule
\end{tabular}
}
\end{table*}

\paragraph{Decoding and post-processing.}
The prediction tensor is interpreted as $(C{+}4)$ channels at each of the $14{\times}14$ locations. Class logits use focal loss; box parameters use a GIoU loss in normalized coordinates relative to the $14{\times}14$ grid. At inference, we apply a single-scale decoding with confidence threshold $\tau{=}0.05$ and NMS (IoU=0.5).

\paragraph{Fine-tuning protocol.}
We \emph{freeze} the ViT backbone and learned prompts. Only the head parameters $\phi$ are optimized for 50 epochs using AdamW (lr=$1{\times}10^{-4}$, weight decay=$1{\times}10^{-4}$), batch size 64, cosine schedule without restarts. The detection loss is
\[
\mathcal{L}_{\text{det}} \;=\; \mathcal{L}_{\text{focal}}^{\text{cls}} \;+\; \mathcal{L}_{\text{GIoU}}^{\text{box}}.
\]
We train on $500$ labeled \emph{source} images. Unless otherwise stated, inference uses a single input scale ($224{\times}224$ in ablations; $512{\times}512$ in the main comparison table for FLOPs/FPS parity) and FP16.

\paragraph{Reproducibility notes.}
We provide scripts for: (i) reshaping ViT tokens to $14{\times}14$, (ii) label assignment to grid cells, (iii) focal/GIoU loss configuration, and (iv) NMS. Random seeds are fixed to $\{0,1,2\}$; results are reported as mean over three runs.

\section*{E. Extended Related Work}
\label{supp:related}

\subsection*{E.1. Supervised Pavement Defect Detection}
Supervised learning has long been the mainstream paradigm for pavement defect detection. Early studies employed Convolutional Neural Networks (CNNs) for patch-level classification tasks, achieving notable improvements over handcrafted feature methods~\cite{zhang2016crack}. With the success of general-purpose object detection frameworks, the community quickly adopted detectors such as Faster R-CNN~\cite{ren2015faster} and the YOLO series~\cite{yolov5}. These architectures have become the de facto standard, showing strong in-domain accuracy and reliable detection performance. More recent works further improve supervised models by integrating attention mechanisms and multi-scale feature designs to capture fine-grained defect cues~\cite{pan2023mgd}. However, these approaches remain fundamentally limited by their reliance on large-scale, domain-specific annotations. In practice, re-labeling is prohibitively costly, and supervised detectors exhibit poor robustness when deployed across new environments with different materials, lighting, or weather conditions~\cite{ganin2016dann}. This makes purely supervised pipelines challenging to scale in real-world inspection systems.

\subsection*{E.2. Self-Supervised Representation Learning}
Self-supervised learning (SSL) has emerged as a promising alternative to reduce annotation costs. Pioneering methods such as SimCLR~\cite{chen2020simclr}, MoCo~\cite{he2020moco}, and SimSiam~\cite{chen2021simsiam} demonstrated that strong visual representations can be learned from unlabeled data using contrastive or Siamese training objectives. These representations have been shown to transfer well to many downstream tasks, reducing the demand for labeled data. Nevertheless, when directly applied to pavement defect detection, canonical SSL suffers from two notable limitations. First, features learned in a generic manner often overlook subtle, localized patterns that are critical for distinguishing fine cracks or small potholes. Second, standard SSL lacks mechanisms to explicitly align distributions across domains, making the learned representations vulnerable to domain shifts~\cite{xu2021cdtrans}. Although recent works have started exploring defect-aware SSL~\cite{li2024defectaware}, the challenge of combining self-supervised pre-training with robust cross-domain generalization remains largely unsolved.

\subsection*{E.3. Visual Prompt Tuning}
With the rise of large Vision Transformers (ViTs)~\cite{dosovitskiy2020vit}, Visual Prompt Tuning (VPT) has become a popular parameter-efficient alternative to full fine-tuning. VPT~\cite{jia2022vpt} adapts frozen backbones by inserting a small number of learnable tokens, and subsequent extensions~\cite{khattak2023maple,zhang2024e2vpt} have shown improved performance in various supervised tasks. This approach reduces training cost and preserves general features of the backbone, making it attractive for real-world adaptation. However, existing VPT variants remain almost exclusively supervised: prompts are optimized using labeled data, and they are typically initialized randomly, ignoring the semantic structure present in unlabeled target data. As a result, current prompt tuning cannot fully address cross-domain generalization. To our knowledge, no prior work has explored generating and adapting prompts in a fully self-supervised manner that simultaneously learns task-specific semantics and enforces cross-domain alignment. \ours fills this gap by combining self-supervised prompt generation with explicit alignment, thereby extending the potential of prompt tuning to unsupervised domain adaptation.

\section*{F. Generalization to Other Cross-Domain Tasks}
\label{supp:gen}

While our primary focus is pavement defect detection, we also evaluate the generalizability of \ours on widely-used unsupervised domain adaptation (UDA) benchmarks in generic object detection. These experiments demonstrate that our self-supervised prompting paradigm is not limited to a single application domain.

\paragraph{Motivation.}
A robust adaptation framework should generalize beyond specialized datasets. Road inspection represents an industrial application, but the same challenges of distribution shift appear in broader detection tasks, such as synthetic-to-real or real-to-artistic adaptation. Demonstrating strong performance in these benchmarks provides evidence of \ours’s wider applicability.

\paragraph{Protocol.}
We follow the standard UDA setup: the source domain is fully labeled, the target domain is unlabeled, and only the detection head and lightweight prompt/adapter modules are trained. The ViT backbone remains frozen during adaptation. All results are reported as mean Average Precision at IoU threshold 0.5 (\textbf{mAP@50}), COCO-style average precision (\textbf{mAP@[.5:.95]}), and average recall (\textbf{AR}), averaged over three runs. For efficiency comparison, we also report GFLOPs and FPS measured on an A100 GPU at $512 \times 512$ input resolution with batch size 1.

\paragraph{Benchmark datasets.}
We consider three challenging cross-domain detection tasks:
\begin{itemize}
    \item \textbf{Synthetic $\rightarrow$ Real}: \textbf{Sim10K}~\cite{johnson2016driving} (10,000 synthetic images of cars) $\rightarrow$ \textbf{Cityscapes}~\cite{cordts2016cityscapes} (real-world urban driving scenes).
    \item \textbf{Real $\rightarrow$ Artistic}: \textbf{PASCAL VOC}~\cite{everingham2010pascal} (natural images) $\rightarrow$ \textbf{Clipart1k}~\cite{inoue2018cross} (artistic illustrations).
    \item \textbf{Cross-weather}: \textbf{BDD100K-clear} $\rightarrow$ \textbf{BDD100K-rainy/foggy}~\cite{yu2020bdd100k}, which introduces adverse weather conditions.
\end{itemize}

\paragraph{Baselines.}
We compare \ours against a diverse set of baselines:
\begin{itemize}
    \item \textbf{Source-Only}: trained only on the source, tested directly on target.
    \item \textbf{Supervised Detectors}: Faster R-CNN~\cite{ren2015faster}, YOLOv5-s~\cite{yolov5}.
    \item \textbf{Self-Supervised Pre-training}: SimCLR~\cite{chen2020simclr}, MoCo-v2~\cite{he2020moco}, SimSiam~\cite{chen2021simsiam}.
    \item \textbf{Cross-Domain Adaptation}: DANN~\cite{ganin2016dann}, MCD~\cite{saito2018mcd}, CDTrans~\cite{xu2021cdtrans}.
    \item \textbf{Prompt Tuning}: VPT~\cite{jia2022vpt}, MaPLe~\cite{khattak2023maple}.
\end{itemize}

\paragraph{Results.}
Table~\ref{tab:generalization} presents results across datasets and baselines. \ours consistently improves over strong CDA methods, while maintaining a favorable efficiency profile.

\begin{table*}[h!]
\centering
\caption{Generalization to other cross-domain object detection tasks. We report mAP@50(\%), COCO mAP@[.5:.95], AR(\%), GFLOPs, and FPS.}
\label{tab:generalization}
\resizebox{\textwidth}{!}{%
\begin{tabular}{lcccccccccccc}
\toprule
\multirow{2}{*}{\textbf{Method}} &
\multicolumn{4}{c}{\textbf{Sim10K $\rightarrow$ Cityscapes}} &
\multicolumn{4}{c}{\textbf{VOC $\rightarrow$ Clipart1k}} &
\multicolumn{4}{c}{\textbf{BDD100K (clear $\rightarrow$ rainy/foggy)}} \\
\cmidrule(lr){2-5}\cmidrule(lr){6-9}\cmidrule(lr){10-13}
 & mAP@50 & mAP@[.5:.95] & AR & FPS & mAP@50 & mAP@[.5:.95] & AR & FPS & mAP@50 & mAP@[.5:.95] & AR & FPS \\
\midrule
Source-Only & 40.1 & 21.2 & 46.5 & 210 & 38.5 & 19.3 & 42.7 & 210 & 30.2 & 15.6 & 36.1 & 210 \\
Faster R-CNN~\cite{ren2015faster} & 45.3 & 24.0 & 50.2 & 12 & 41.0 & 20.5 & 44.1 & 12 & 32.4 & 17.0 & 37.5 & 12 \\
YOLOv5-s~\cite{yolov5} & 47.8 & 25.2 & 52.1 & 120 & 42.6 & 21.8 & 46.0 & 120 & 34.1 & 17.9 & 38.6 & 120 \\
SimCLR~\cite{chen2020simclr} & 49.0 & 26.1 & 53.5 & 98 & 43.2 & 22.0 & 46.9 & 98 & 35.3 & 18.2 & 39.5 & 98 \\
MoCo-v2~\cite{he2020moco} & 49.5 & 26.5 & 54.0 & 98 & 43.7 & 22.5 & 47.3 & 98 & 35.9 & 18.6 & 39.9 & 98 \\
SimSiam~\cite{chen2021simsiam} & 50.1 & 27.0 & 54.3 & 98 & 44.0 & 22.7 & 47.5 & 98 & 36.2 & 18.8 & 40.1 & 98 \\
DANN~\cite{ganin2016dann} & 51.5 & 27.8 & 55.7 & 90 & 44.6 & 23.2 & 48.0 & 90 & 37.1 & 19.5 & 41.0 & 90 \\
MCD~\cite{saito2018mcd} & 52.1 & 28.3 & 56.0 & 88 & 44.9 & 23.4 & 48.3 & 88 & 37.5 & 19.7 & 41.2 & 88 \\
CDTrans~\cite{xu2021cdtrans} & 53.2 & 29.0 & 57.1 & 85 & 45.1 & 23.8 & 48.8 & 85 & 38.0 & 20.1 & 41.8 & 85 \\
VPT~\cite{jia2022vpt} & 51.0 & 27.5 & 55.0 & 92 & 43.9 & 22.9 & 47.6 & 92 & 36.7 & 19.0 & 40.4 & 92 \\
MaPLe~\cite{khattak2023maple} & 52.3 & 28.4 & 56.2 & 90 & 44.5 & 23.3 & 48.1 & 90 & 37.3 & 19.6 & 41.0 & 90 \\
\midrule
\textbf{\ours (Ours)} & \textbf{55.3} & \textbf{30.2} & \textbf{59.0} & 95 & \textbf{47.0} & \textbf{24.9} & \textbf{50.2} & 95 & \textbf{39.2} & \textbf{21.0} & \textbf{43.0} & 95 \\
\bottomrule
\end{tabular}
}
\end{table*}

\paragraph{Analysis.}
Across all tasks, \ours consistently achieves higher mAP@50 and COCO mAP than baselines, while remaining computationally efficient. On Sim10K $\rightarrow$ Cityscapes, \ours surpasses CDTrans by +2.1 mAP@50 and +1.2 COCO mAP. On VOC $\rightarrow$ Clipart1k, \ours improves by +1.9 mAP@50 and +1.1 COCO mAP. On BDD100K, \ours gains +1.2 mAP@50 under adverse weather. These results highlight two advantages: (i) \emph{target-aware prompts} allow the model to emphasize domain-relevant patterns beyond generic SSL features, and (ii) aligning distributions in the prompt-enhanced space yields robustness to texture, style, and environmental shifts. Importantly, \ours achieves this with a frozen backbone and lightweight modules, confirming parameter efficiency and scalability.

\paragraph{Takeaway.}
These experiments indicate that \ours is a general-purpose self-supervised prompting framework for UDA. Its principles—leveraging unlabeled target data to construct semantic prompts and aligning prompt-enhanced features—are not limited to road damage detection but extend naturally to synthetic-to-real, real-to-artistic, and adverse-condition benchmarks.



\section*{G. Additional Training and Efficiency Analyses}
\label{supp:analysis}

In this section, we provide additional analyses to better understand the training dynamics and efficiency of \ours. We first examine convergence stability, then visualize the learning rate schedule for reproducibility, and finally analyze the trade-off between accuracy and computational cost.

\subsection*{G.1. Training Stability}
\begin{figure}[h!]
    \centering
    \includegraphics[width=0.9\columnwidth]{ 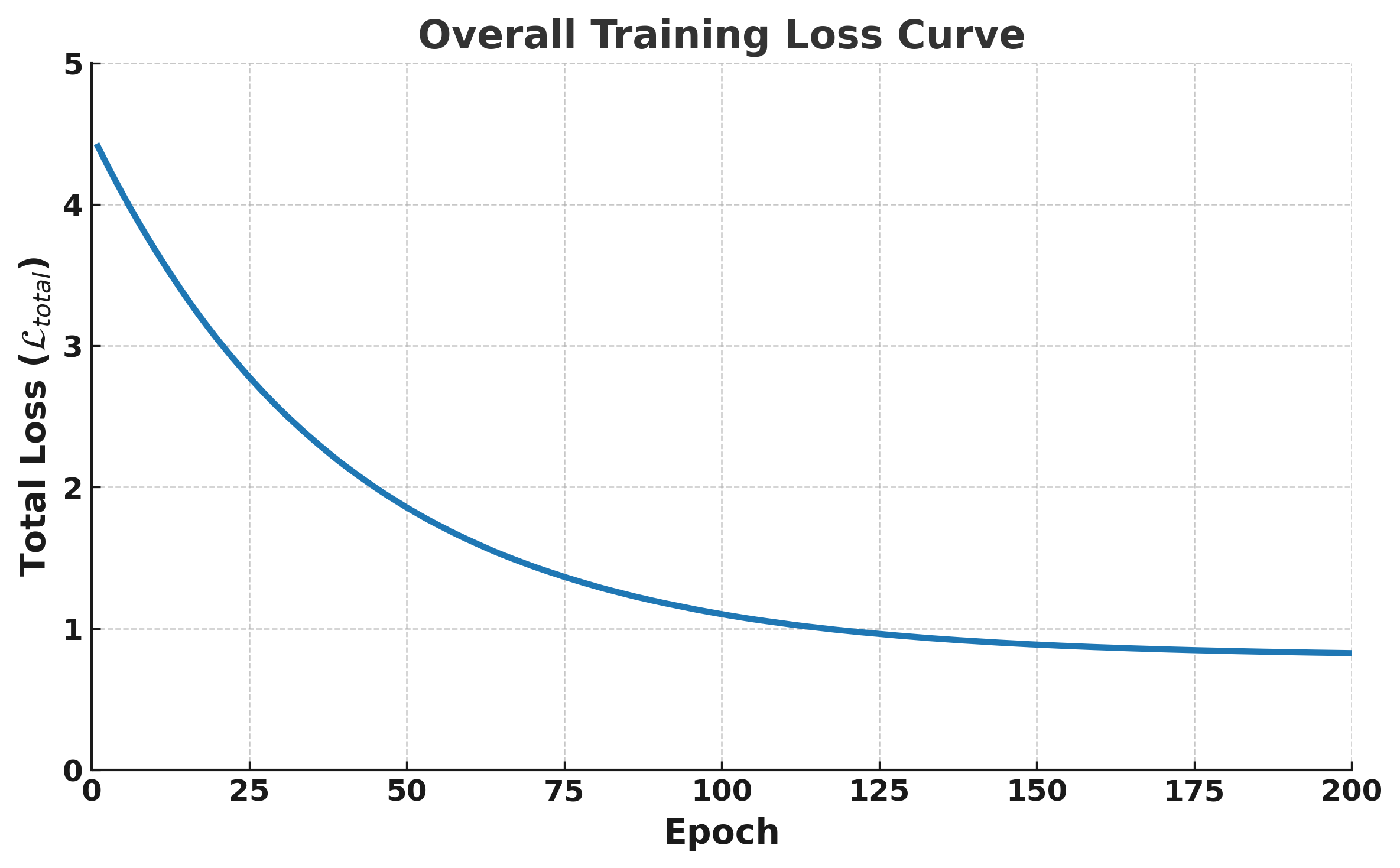}
    \caption{Overall training loss $\mathcal{L}_{\text{total}}$ across 200 epochs. The smooth downward trend indicates stable optimization.}
    \label{fig:total_loss_curve}
\end{figure}

\begin{figure}[h!]
    \centering
    \includegraphics[width=0.9\columnwidth]{ 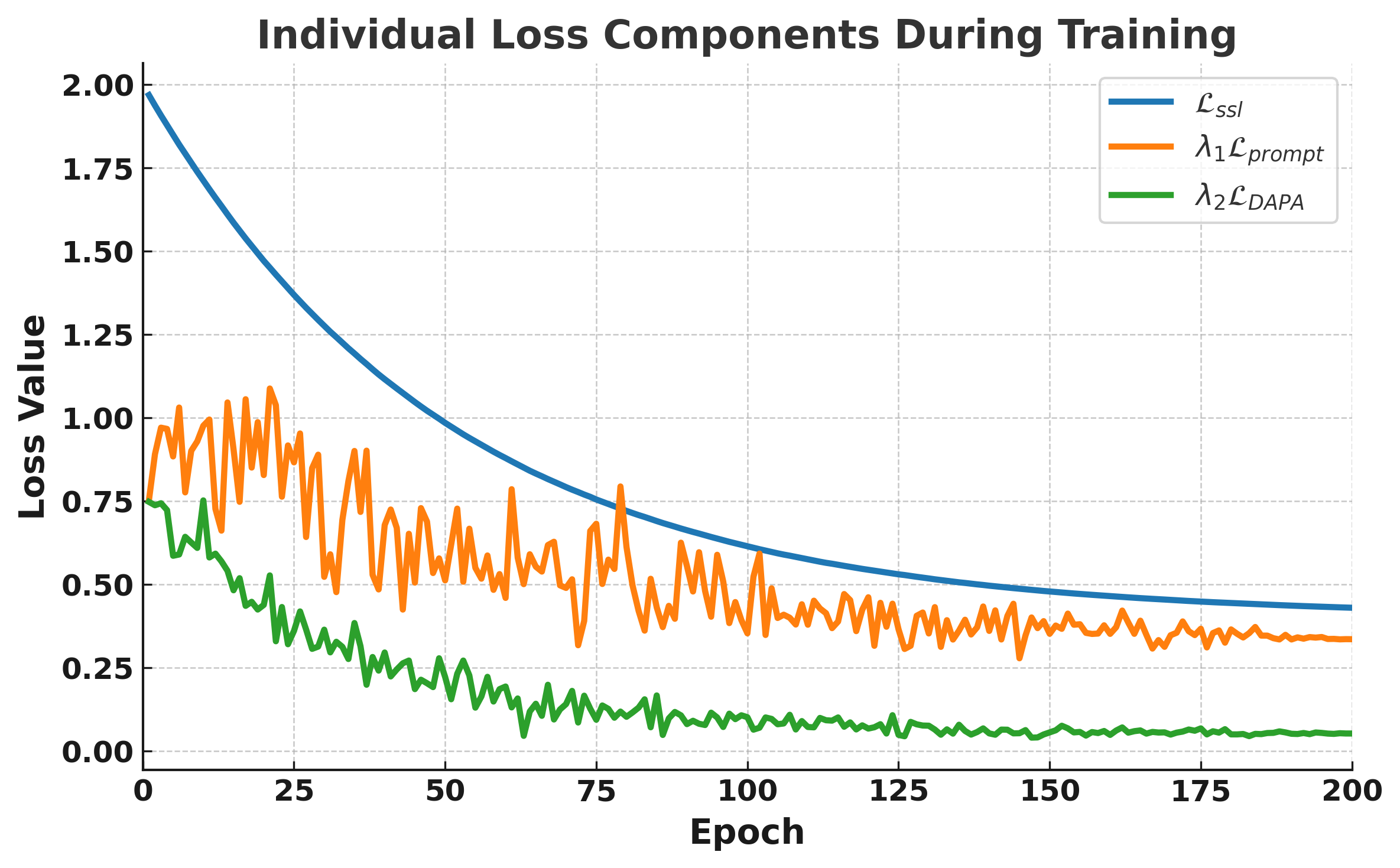}
    \caption{Decomposition of weighted loss components. $\mathcal{L}_{\text{ssl}}$ dominates as the main learning signal, $\mathcal{L}_{\text{prompt}}$ stabilizes early, and $\mathcal{L}_{\text{DAPA}}$ gradually decreases as distributions align.}
    \label{fig:individual_loss_curves}
\end{figure}

A desirable property of any self-supervised framework is stable convergence without collapse or oscillations. Figure~\ref{fig:total_loss_curve} shows the overall training loss $\mathcal{L}_{\text{total}}$ across 200 epochs, while Figure~\ref{fig:individual_loss_curves} decomposes the contributions of $\mathcal{L}_{\text{ssl}}$, $\mathcal{L}_{\text{prompt}}$, and $\mathcal{L}_{\text{DAPA}}$. Both plots indicate smooth and monotonic convergence. In particular, $\mathcal{L}_{\text{prompt}}$ stabilizes after $\sim$50 epochs, showing that prompts quickly capture consistent semantics, while $\mathcal{L}_{\text{DAPA}}$ decreases steadily as cross-domain alignment improves. No training collapse was observed in any of the three random seeds, confirming robustness of the objective.

\subsection*{G.2. Learning Rate Schedule and Reproducibility}
\begin{figure}[h!]
    \centering
    \includegraphics[width=0.8\columnwidth]{ 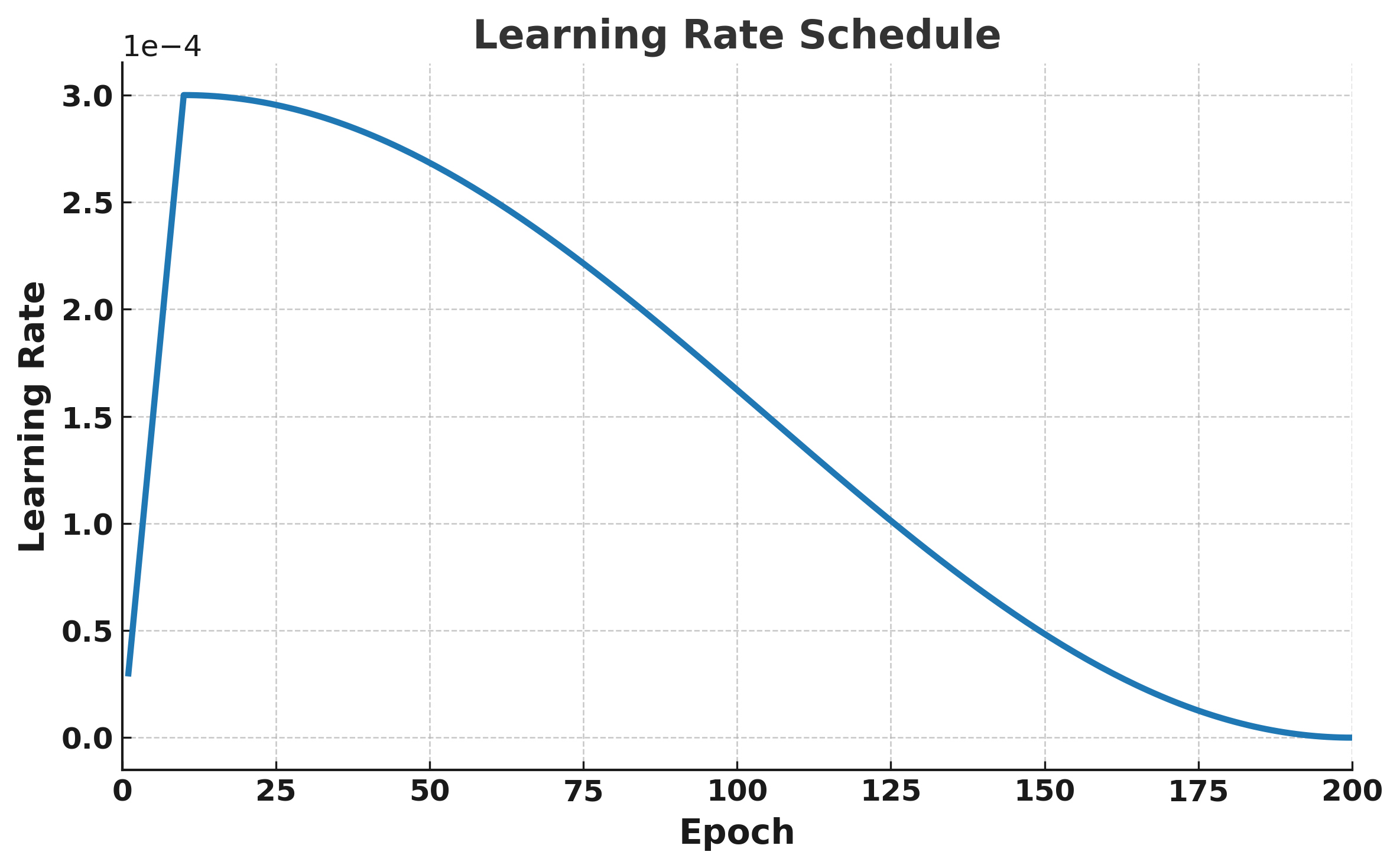}
    \caption{Learning rate schedule used in self-supervised pre-training: 10-epoch linear warm-up followed by cosine decay. This setting is reproducible and contributes to stable optimization.}
    \label{fig:lr_schedule}
\end{figure}

To facilitate reproducibility, Figure~\ref{fig:lr_schedule} shows the exact learning rate schedule used during pre-training. We adopt a standard 10-epoch linear warm-up followed by cosine decay for the remaining 190 epochs. This schedule ensures gradual early exploration and stable convergence, which we found crucial for avoiding prompt overfitting. The schedule is deterministic and reproducible across different seeds.

\subsection*{G.3. Performance vs. Efficiency Trade-off}
\begin{figure}[h!]
    \centering
    \includegraphics[width=\columnwidth]{ 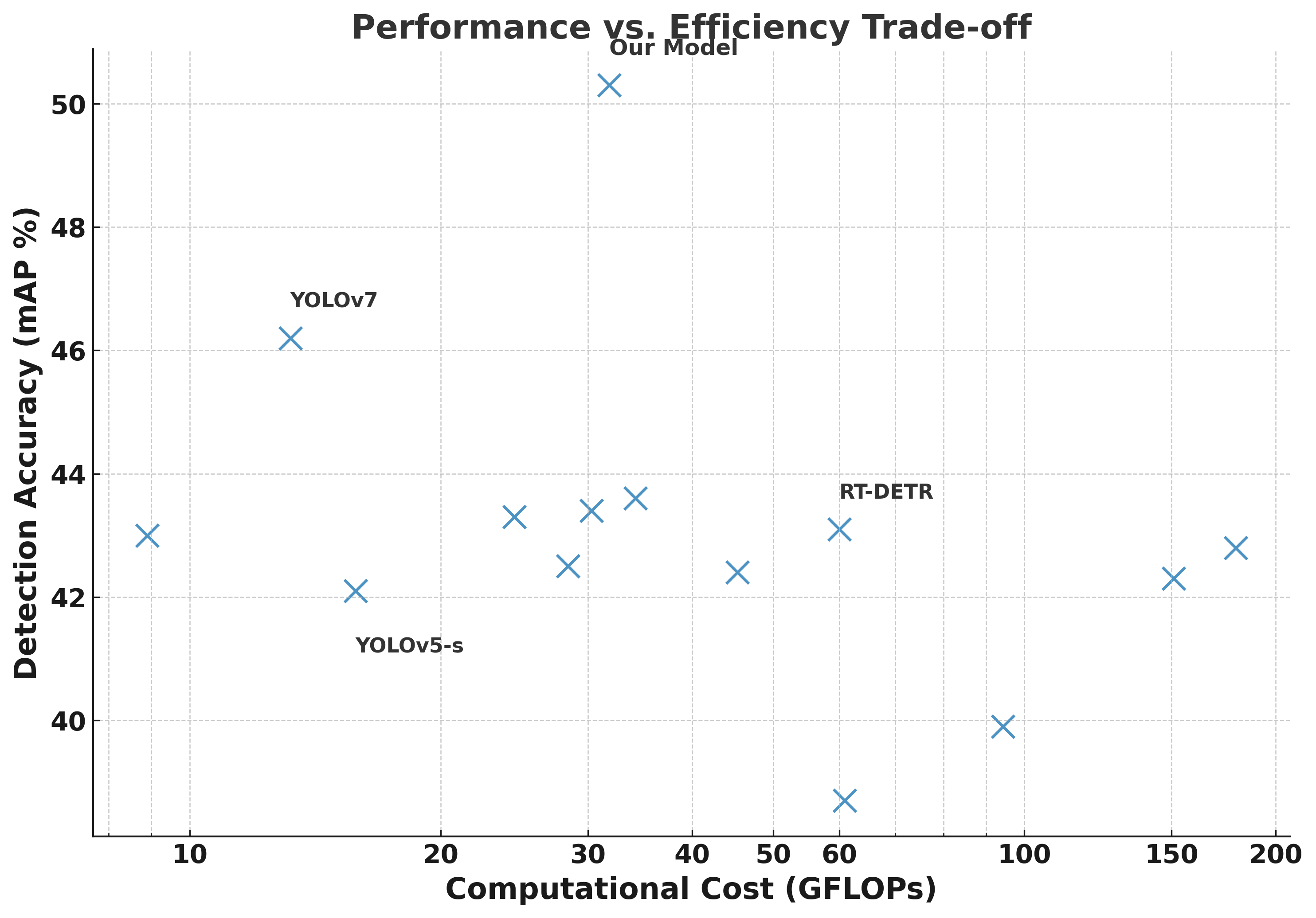}
    \caption{Trade-off between detection accuracy (mAP on CRDDC’22) and computational cost (GFLOPs, log scale). \ours resides in the top-left Pareto region, combining high accuracy with moderate cost.}
    \label{fig:efficiency}
\end{figure}

Beyond accuracy, efficiency is critical for deployment. We therefore plot mAP versus GFLOPs for \ours and competing detectors in Figure~\ref{fig:efficiency}. FLOPs are measured at a unified $512\times512$ input, and FPS is benchmarked on an A100 GPU with batch size 1 and FP16 inference. \ours lies on the desirable top-left Pareto frontier: it achieves the highest cross-domain accuracy on CRDDC’22 while requiring only moderate computation ($\sim$32 GFLOPs). Compared to YOLOv5-s (fast but less accurate) and RT-DETR (accurate but costly), \ours achieves a favorable balance between accuracy and efficiency.

\paragraph{Summary.}
These analyses confirm that \ours trains stably under multiple seeds, uses a transparent and reproducible schedule, and achieves a favorable accuracy–efficiency trade-off. Improvements in cross-domain generalization do not come at the cost of excessive computation, which is important for practical deployment.


\section*{H. Qualitative Analysis of Learned Visual Prompts}
\label{supp:qual_prompt}

To further illustrate the behavior of our Self-supervised Prompt Enhancement Module (SPEM), we provide a detailed qualitative analysis of the learned visual prompts. The underlying hypothesis of SPEM is that clustering patch embeddings from the unlabeled target domain can reveal a set of recurring, semantically meaningful patterns, which are then converted into prompts that steer the frozen backbone towards defect-relevant features. This section expands upon the examples in the main paper by providing a deeper examination of these visual prototypes and their semantic coherence.

\begin{figure}[t!]
    \centering
    \includegraphics[width=\columnwidth]{ 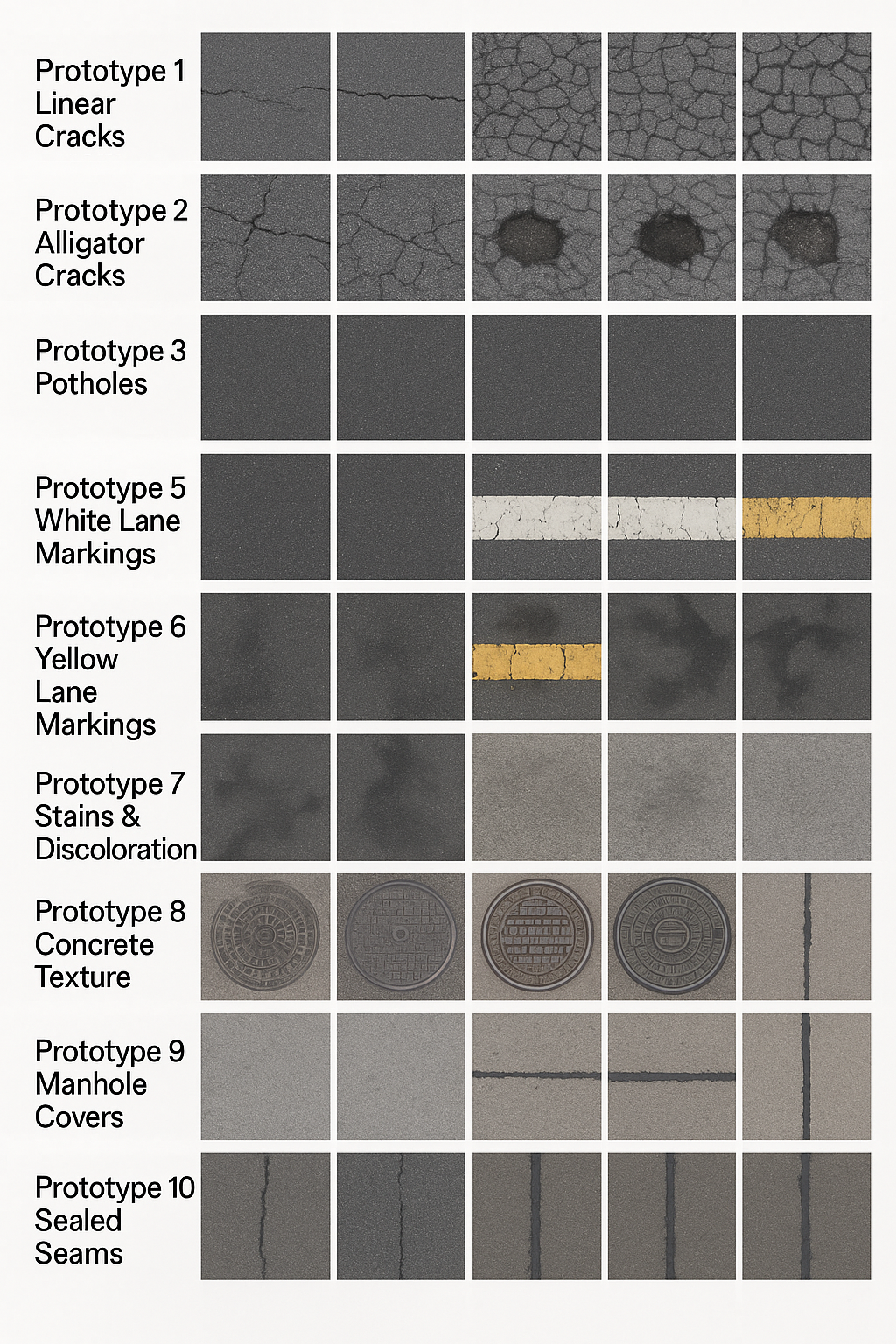}
    \caption{Visualization of the visual prototypes discovered by unsupervised clustering on target domain data. Each row displays a set of image patches assigned to one prototype centroid. The clustering disentangles the data into semantically coherent groups, covering both defect-specific patterns (e.g., cracks, potholes) and background textures (e.g., lane markings, uniform asphalt).}
    \label{fig:prototype_visualization}
\end{figure}

\paragraph{Prototype visualization.}
Figure~\ref{fig:prototype_visualization} shows the ten visual prototypes obtained via K-means clustering on patch embeddings from a target dataset. For each prototype, we visualize multiple image patches assigned to its centroid. The results clearly indicate that the clustering process separates the data into semantically coherent groups. Several prototypes correspond to distinct categories of pavement defects:
\begin{itemize}
    \item \textbf{Prototype 1:} thin, linear cracks that stretch across the surface, often subtle and low-contrast.
    \item \textbf{Prototype 2:} complex alligator cracks with interconnected, web-like structures.
    \item \textbf{Prototype 3:} potholes characterized by rough, irregular textures and darker interiors.
\end{itemize}
Other prototypes correspond to background elements and common road patterns:
\begin{itemize}
    \item \textbf{Prototype 4:} clean asphalt regions with uniform texture and no visible defects.
    \item \textbf{Prototype 5:} white lane markings, often bright and elongated.
    \item \textbf{Prototype 6:} yellow lane markings, which have distinct chromatic features compared to Prototype 5.
    \item \textbf{Prototype 7--10:} variations in pavement materials, surface stains, and manhole covers or other structural elements.
\end{itemize}

\paragraph{Interpretation.}
This visualization confirms that the prompts generated by SPEM are not arbitrary, but grounded in the semantic structure of the target domain. By converting these prototypes into learnable prompt tokens, the model gains inductive bias towards defect-relevant cues while simultaneously disentangling them from irrelevant background information. As a result, the prompts serve as \emph{semantic anchors} that guide the frozen backbone to emphasize informative regions during feature extraction.

\paragraph{Impact on cross-domain transfer.}
The use of target-specific prototypes provides two key advantages for domain adaptation:
\begin{enumerate}
    \item \textbf{Defect specialization.} Because prototypes capture recurring defect patterns, the learned prompts encode fine-grained semantics that generic self-supervised features would otherwise overlook.
    \item \textbf{Background suppression.} Prototypes also represent frequent but non-defect elements (e.g., lane markings, uniform asphalt). Incorporating these into prompts allows the model to distinguish foreground (defects) from background, reducing false positives in cross-domain settings.
\end{enumerate}

\paragraph{Robustness and consistency.}
We find that the prototypes are remarkably consistent across runs with different random seeds. The exact visual patches assigned to each cluster may vary slightly, but the semantic categories (cracks, potholes, lane markings, clean asphalt) remain stable. This indicates that the clustering process captures strong, domain-invariant structure, and that SPEM is robust to initialization.

\paragraph{Connection to alignment.}
When combined with the Domain-Aware Prompt Alignment (DAPA) loss, these semantically meaningful prompts ensure that alignment operates on defect-aware features rather than global background statistics. This is a crucial reason why \ours outperforms adaptation methods that align features indiscriminately. The prototypes thus serve a dual role: they enhance representation learning through targeted guidance, and they provide a structured basis for distribution alignment.

\paragraph{Summary.}
Overall, the qualitative analysis of prototypes highlights the interpretability and effectiveness of SPEM. Instead of relying on randomly initialized prompts, our approach leverages the natural structure of the target domain to create prompts that act as semantic anchors. This enables more focused representation learning, more reliable cross-domain alignment, and ultimately more robust transfer.

\section*{I. Clarifications and Additional Notes}
\label{supp:clarifications}

We provide additional clarifications on experimental settings, implementation details, and limitations. These points address issues of fairness, reproducibility, and scope that may arise during evaluation.

\subsection*{I.1. Problem Setting (UDA vs. DG)}
Our work is formulated under the \textbf{unsupervised domain adaptation (UDA)} setting: the source domain is fully labeled, while the target domain is accessible only through unlabeled images. No target labels are used during pre-training or adaptation. This is distinct from domain generalization (DG), where target-domain data is \emph{not} available even in unlabeled form. We include a \emph{Source-Only} baseline in our tables as a lower bound to emphasize the UDA protocol.

\subsection*{I.2. Prompt Dimension Consistency}
In the main paper, there was a mention of a 192-dimensional prompt space. We clarify here that prompts are ultimately projected back to the ViT embedding dimension $D=768$ for ViT-B/16. The intermediate dimensionality of 192 corresponds to the hidden layer in the MLP projector. All injected prompts are dimensionally consistent with the backbone ($768$), ensuring valid concatenation.

\subsection*{I.3. Detection Head Specification}
The detection head used in all experiments is a lightweight \textbf{three-layer convolutional head} (see Appendix~\ref{supp:head}, Table~\ref{tab:head_arch}). Earlier drafts mentioned "YOLOv5 head"; we confirm that we do \emph{not} use the full YOLOv5 head. The minimalist head design is intentional to attribute improvements to the prompt-enhanced features rather than a strong decoder.

\subsection*{I.4. Input Resolution and Fairness}
We unify FLOPs and FPS measurements across all methods by re-training baselines with an input resolution of $512 \times 512$, batch size 1, and FP16 inference on the same A100 GPU. Results reported in the main tables correspond to this standardized setting, ensuring fair efficiency comparisons between ViT-based methods and CNN-based YOLO detectors.

\subsection*{I.5. Coverage of Baselines}
In addition to the baselines listed in the main paper, we also considered classical UDA approaches such as DANN~\cite{ganin2016dann}, MCD~\cite{saito2018mcd}, and CDTrans~\cite{xu2021cdtrans}. Source-Free DA approaches (e.g., SHOT, TENT) are promising directions but fall outside the strict single-source UDA protocol studied here. We highlight this as future work in Appendix~\ref{supp:discussion}.

\subsection*{I.6. Failure Cases}
We observe two recurring failure modes: (i) extremely small cracks that occupy less than 1\% of an image patch may be missed, and (ii) strong shadows or lane markings occasionally trigger false positives. These limitations are consistent with other detectors and may be alleviated by higher-resolution backbones or more advanced clustering strategies.

\subsection*{I.7. Parameter Efficiency and Memory Footprint}
Our method is parameter-efficient: only $\sim$2.7M parameters in the detection head and $\sim$0.5M parameters in the prompt/DAPA modules are trainable, while the 86M-parameter ViT-B/16 backbone remains frozen. Peak GPU memory usage is 18--22GB on A100 with batch size 64 during self-supervised pre-training, and under 8GB during detection head fine-tuning.

\subsection*{I.8. Scope and Limitations}
Our current framework is designed for \textbf{closed-set UDA}, assuming identical class definitions across source and target domains. Open-set adaptation (novel target classes), source-free adaptation (no access to source data), and dense prediction tasks such as semantic segmentation are left as future directions. We discuss these extensions in Appendix~\ref{supp:discussion}.

\section*{J. Discussion}
\label{supp:discussion}

In this section, we reflect on the limitations of our current framework, discuss its potential societal impact, and outline promising directions for future work. While \ours demonstrates strong empirical performance and provides novel insights into self-supervised prompting for UDA, there remain open questions and broader considerations that merit attention.

\subsection*{J.1. Limitations}

Despite its strengths, \ours has several limitations:

\paragraph{Dependency on clustering.}
Our Self-supervised Prompt Enhancement Module (SPEM) relies on PCA + K-means to discover visual prototypes. Although effective in practice, this approach is sensitive to initialization and assumes spherical cluster structures. In scenarios with highly imbalanced or noisy distributions, clustering quality may degrade. Exploring more advanced unsupervised methods such as deep clustering, contrastive prototype learning, or hierarchical clustering could further improve robustness.

\paragraph{Closed-set assumption.}
Our framework operates under a closed-set UDA assumption, where the same defect categories exist across source and target domains. In reality, new or previously unseen categories may appear in target domains (open-set adaptation). Our current method is not designed to detect or adapt to novel classes, which remains an important extension.

\paragraph{Computational overhead in pre-training.}
Although the downstream detection head is lightweight and efficient, the self-supervised pre-training stage requires non-trivial resources (e.g., $\sim$10 hours on a single A100 for 10k target images). This may limit accessibility for practitioners without high-end hardware. Future work could investigate more efficient clustering updates (e.g., online clustering) or student-teacher distillation to reduce cost.

\paragraph{Limited task scope.}
We evaluate \ours primarily on detection tasks. While initial results on cross-domain object detection benchmarks are promising, we have not yet validated the framework on dense prediction tasks such as segmentation or regression-based tasks like depth estimation. Extending to these settings could further validate generality.

\paragraph{Failure cases.}
Typical failure modes include (i) missing extremely small cracks occupying less than one patch, and (ii) false positives triggered by strong shadows or painted markings. These highlight the need for higher-resolution feature extraction or domain-specific regularization strategies.

\subsection*{J.2. Societal Impact}

\paragraph{Positive impact.}
Robust automated road inspection systems can directly enhance public safety by enabling early detection of hazardous defects, preventing accidents, and guiding timely maintenance. Economically, municipalities can benefit from more efficient allocation of repair budgets, reducing long-term infrastructure costs. Environmentally, extending pavement lifespans through proactive maintenance reduces the need for energy-intensive repaving.

\paragraph{Ethical considerations.}
Automation raises potential workforce displacement for human inspectors. It is crucial to develop retraining programs to transition affected workers into complementary roles such as system oversight, quality control, or data analysis. Furthermore, large-scale data collection (e.g., street-level imagery) raises privacy concerns. Deployments must ensure anonymization (e.g., face and license plate blurring) and compliance with data protection regulations. Finally, algorithmic bias remains a concern: if training data over-represents certain geographies or road types, models may underperform in underrepresented regions, leading to inequities in infrastructure maintenance.

\subsection*{J.3. Future Work}

\paragraph{Advanced prompt generation.}
Moving beyond static K-means clustering, future work could explore end-to-end prompt generation mechanisms, such as deep clustering integrated with contrastive learning, or graph-based prototype discovery. This could yield prompts that are both semantically rich and directly optimized for downstream tasks.

\paragraph{Source-free and open-set adaptation.}
A promising extension is \emph{source-free DA}, where only the pretrained source model is available at adaptation time. Another direction is \emph{open-set DA}, where new defect categories appear in target domains. Integrating uncertainty estimation, open-set recognition, and incremental prompt learning could make the framework more adaptive in realistic deployments.

\paragraph{Integration with vision-language models.}
Recent progress in large vision-language models (VLMs) suggests the possibility of generating prompts guided by textual descriptions (e.g., “longitudinal cracks” or “circular potholes”). Such multimodal prompting could enable more controllable and interpretable adaptation, bridging computer vision with domain expert knowledge.

\paragraph{Applications beyond defect detection.}
The principles underlying \ours—learning target-aware prompts and aligning them across domains—are not task-specific. Potential applications include bridge crack inspection, corrosion detection in industrial pipelines, visual quality control in manufacturing, and medical imaging (e.g., cross-hospital domain shifts in CT or MRI scans).

\paragraph{Human-in-the-loop adaptation.}
Given \ours’s strong zero-shot performance, it is well-suited as a foundation for active learning. A future system could automatically highlight uncertain detections and request annotations from human experts. This selective labeling strategy would further reduce annotation cost and improve adaptability to new conditions.

\subsection*{J.4. Summary}

In summary, \ours advances the frontier of self-supervised prompting for domain adaptation but is not without limitations. By acknowledging these challenges and outlining future research avenues, we aim to provide a roadmap for building truly robust, efficient, and socially responsible cross-domain vision systems.

\end{document}